\documentclass[twoside,11pt]{article}

\usepackage{blindtext}

\usepackage{jmlr2e}

\usepackage[utf8]{inputenc} 
\usepackage[T1]{fontenc}    
\usepackage{url}            
\usepackage{amsfonts}       
\usepackage{nicefrac}       
\usepackage{microtype}      
\usepackage{xcolor}         

\usepackage{microtype}
\usepackage{graphicx}
\usepackage{subfigure}
\usepackage{adjustbox}
\usepackage{booktabs} 
\usepackage[leftmargin=3em]{quoting}
\usepackage{caption}

\usepackage{amssymb}
\usepackage{amsmath}
\usepackage{algpseudocode}
\usepackage{algorithm} 
\usepackage{multirow,multicol,ctable,tabularx} 
\allowdisplaybreaks
\usepackage{hyperref}
\usepackage{mathtools}
\usepackage{wrapfig}

\usepackage{adjustbox}
\usepackage{booktabs} 
\usepackage{textcomp}
\usepackage{amsmath}
\usepackage{threeparttable}
\usepackage{dsfont}
\usepackage{algorithm}
\usepackage{amssymb}

\hypersetup{hidelinks}

\newtheorem{theorem}{Theorem}

\newtheorem{assumption}{Assumption}
\newtheorem{corollary}{Corollary}
\newtheorem{appendixcorollary}{Corollary}

\newtheorem{claim}{Claim}
\newtheorem{appendixclaim}{Claim}
\newtheorem{definition}{Definition}
\newtheorem{appendixdefinition}{Definition}

\usepackage{lastpage}
\jmlrheading{25}{2024}{1-\pageref{LastPage}}{3/23; Revised 12/23}{1/24}{23-0314}{Dahuin Jung, Hyungyu Lee, and Sungroh Yoon}
\ShortHeadings{Sample-efficient Adversarial Imitation Learning}{Jung, Lee, and Yoon}

\firstpageno{1}

\begin{document}

\title{Sample-efficient Adversarial Imitation Learning}

\author{
\name \hspace{-0.33em}Dahuin Jung
\email annajung0625@snu.ac.kr \\
\addr Electrical and Computer Engineering \\
Seoul National University, Seoul 08826, Republic of Korea \\
\name Hyungyu Lee  
\email rucy74@snu.ac.kr \\
\addr Electrical and Computer Engineering \\
Seoul National University, Seoul 08826, Republic of Korea \\
\name Sungroh Yoon\thanks{Corresponding Author \\A preliminary version of this manuscript was presented at Deep RL Workshop, NeurIPS 2022.}
\email sryoon@snu.ac.kr \\
\addr Electrical and Computer Engineering \\
Interdisciplinary Program in Artificial Intelligence \\ 
Seoul National University, Seoul 08826, Republic of Korea \\
}

\editor{Scott Niekum}

\maketitle

\begin{abstract}

Imitation learning, in which learning is performed by demonstration, has been studied and advanced for sequential decision-making tasks in which a reward function is not predefined. However, imitation learning methods still require numerous expert demonstration samples to successfully imitate an expert's behavior.~To improve sample efficiency, we utilize self-supervised representation learning, which can generate vast training signals from the given data. In this study, we propose a self-supervised representation-based adversarial imitation learning method to learn state and action representations that are robust to diverse distortions and temporally predictive, on non-image control tasks. In particular, in comparison with existing self-supervised learning methods for tabular data, we propose a different corruption method for state and action representations that is robust to diverse distortions. We theoretically and empirically observe that making an informative feature manifold with less sample complexity significantly improves the performance of imitation learning. The proposed method shows a 39$\%$ relative improvement over existing adversarial imitation learning methods on MuJoCo in a setting limited to 100 expert state-action pairs. Moreover, we conduct comprehensive ablations and additional experiments using demonstrations with varying optimality to provide insights into a range of factors.

\end{abstract}

\begin{keywords}
  imitation learning, adversarial imitation learning, self-supervised learning, data efficiency
\end{keywords}
\section{Introduction}

Imitation learning (IL) is widely used in sequential decision-making tasks, where the design of a reward function is complicated or uncertain. When a reward is sparse~\citep{sqil} or an optimal reward function is unknown, IL finds an optimal policy that relies only on expert demonstrations. Owing to recent development in deep neural networks, the range of behaviors, which can be imitated, has expanded. There are two main learning approaches for IL. The first approach trains a policy by following actions from an expert in a supervised manner called behavioral cloning (BC)~\citep{bc_original,alvinn}. However, error accumulation limits BC because it greedily imitates the demonstrated actions. For behavior cloning, it talks about the problem of accumulating errors, but there are alternative IL approaches like DAgger that deals with that. The second approach is inverse reinforcement learning (IRL)~\citep{irl_survey}, inferring a cost function based on given expert demonstrations. The IRL implements adversarial learning~\citep{gan} to infer the cost function. Therefore, an agent learns the policy to imitate expert demonstrations, whereas a discriminator learns to differentiate between the expert's behavior and that of the agent. The learned discriminator is used as the cost function in the reinforcement learning (RL) phase. \textcolor{black}{However, there are also several IRL methods that do not leverage adversarial training, e.g., Max-Margin Planning~\citep{maxmargin} and Max-Ent IRL~\citep{maxent}.}

Although IRL has led an advance in IL, it has key challenges. First, adversarial learning is known to be delicate in practice. The min-max computational formulation of adversarial imitation learning (AIL) often involves brittle approximation techniques. Second, the IL requires many demonstration trajectories to recover an expert policy. Although IRL requires fewer demonstrations than BC, it still requires considerable trajectories. Recently, many algorithms or techniques have been proposed to address the first challenge~\citep{airl,vail,sqil,eairl,computation,stablegail}; however, little work has been done to improve the sample efficiency of expert demonstrations required~\citep{fgail,asaf}.

Self-supervised representation learning (SSL) has advanced sample efficiency in the image and language domains~\citep{contrastive:simclr,byol,wordembedding}. It applies various transformations to the data and uses the transformed data itself as supervision. Thus, it increases the sample efficiency by obtaining training signals from auxiliary tasks or objectives that do not rely on labels. Specifically, InfoNCE~\citep{contrastive:simclr,contrastive:moco} and asymmetric twin-based~\citep{byol,simsiam} SSL approaches are known to be effective for learning robust feature representations for different distortions of identical inputs. Recently, SSL has been utilized in image-based RL algorithms~\citep{contrastive:curl,selfpredictive,sslrl} and has shown significant improvement in performance. However, transformation techniques applied to image-based RL are not directly adaptable to non-image control benchmarks. This is because these approaches rely upon the semantic/spatial properties of data that may generate either out-of-distribution examples or examples that supply only the same view when directly applied to a continuous control (tabular) domain.

In this study, we propose a sample-efficient AIL method for non-image control benchmarks. The proposed method leverages auxiliary training signals for learning state and action representations that are temporally predictive and robust to diverse distortions. Based on the characteristics of each domain and benchmark, an auxiliary task that can learn informative feature representations is different. For RL, to address sequential decision-making tasks, the feature representation of a state and action should contain temporally predictive information. \textcolor{black}{To address this, we add an auxiliary task that predicts the next state representation from the given current state and action representations. This marks the initial attempt to enhance the sample efficiency of expert demonstrations in non-image imitation learning by employing SSL.}

Moreover, learning representations that discard information regarding nuisance variables improves generalization and decreases required sample complexity. Previous transformation techniques for tabular data~\citep{neurips2020vime,iclr2022scarf} generate transformed samples far from real samples. Therefore, we propose a simple, effective corruption method that generates transformed samples showing diverse distortions that are possible in-distribution. Empirically, we demonstrate that promoting temporally predictive feature representations with robustness against diverse distortions significantly improves sample efficiency.

First, we theoretically observe that IL performs substantially better when an informative feature manifold is created with less sample complexity. We evaluate the proposed method on MuJoCo~\citep{mujoco} and Atari RAM of OpenAI Gym~\citep{openai}, where each benchmark is allowed less than 100 expert state-action pairs. The proposed method outperforms the previous AIL methods by a significant margin in scenarios with a small number of perfect or imperfect expert demonstrations. We discover that the proposed corruption method generates various transformed samples that are not out-of-distribution. We perform comprehensive ablation studies, detailing the intuitions and effects of various design choices and factors. Through runs with various hyperparameters on non-image control tasks, we demonstrate that the corruption rate and loss function should be carefully chosen.

\section{Related Work}
\subsection{Data-efficient Reinforcement Learning}

In deep RL, studies have been conducted to improve sample efficiency. For continuous control, several studies have suggested the use of reconstruction loss \citep{controlrecon,controlrecon2}. However, most of the suggested methods are RL benchmarks, which have a sparse reward or image state. Methods using a self-supervised error as an intrinsic reward have been proposed to improve the sample efficiency in a sparse reward scenario \citep{curiosity,information,reward_exploration,intrinsicreward2}. For the image state, various image augmentation techniques and self-supervised objectives have been applied to reduce environmental interactions~ \citep{contrastive:curl,selfpredictive,efficiency1,efficiency2,efficiency3,efficiency4,ssp,stabilizing,sslrlimage}. \textcolor{black}{Additionally, there are some works that apply self-supervised learning to train latent space or feature embedding for reward learning~\citep{brown2020safe,bobu2023sirl}.} Different from most existing approaches, we utilize SSL to improve the sample efficiency of expert demonstrations for non-image imitation learning.


\subsection{Self-supervised Representation Learning}

Currently, SSL is divided into three approaches. The first is a pretext task~\citep{jigsaw}, which creates a pre-task that can learn useful feature representations and use the learned representations on downstream tasks. The second is UCL. Contrastive learning works on a simple push-pull principle, and it can be a sample or cluster level~\citep{swav_cluster}.~The contrastive loss contrasts the neighboring instances with non-neighboring ones \citep{contrastive:simclr,contrastive:moco}. The third is asymmetric twin-based SSL. Unlike UCL, these methods do not use negative samples during training. Asymmetric twin-based SSL methods learn robust representations in such a way that differently transformed versions of input have the same representation. As representative methods, BYOL~\citep{byol} and Simsiam~\citep{simsiam} used Siamese networks with weight sharing and stop-gradient techniques to avoid collapse. Barlow twins~\citep{barlow} utilized a correlation matrix between the representations of a differently transformed same input to maximize the similarity between them while minimizing redundancy. VICReg~\citep{vicreg} is effective for making the two representations similar and reducing the embedding of non-informative factors.


There are two SSL methods for tabular data, VIME~\citep{neurips2020vime} and SCARF \citep{iclr2022scarf}, which can be directly applied to non-image control benchmarks. VIME uses a random corruption method and SCARF suggests a method that replaces each feature dimension by a random draw from that feature dimension's empirical marginal distribution. The difference between our work is the corrupted data of swapping is a mixture of only two state-action pairs because the replaced features are sampled from another single state-action pair. Meanwhile, the corrupted data of SCARF is a mixture of varying state-action pairs, resulting in possible out-of-distribution data. In the experiments, the proposed swapping corruption method showed higher performance compared to the other two methods. We measured the variance and outlier scores of the corrupted samples produced by VIME, SCARF, and the proposed method, to confirm that the proposed method qualitatively generates more realistic and still varied samples.



\subsection{Inverse Reinforcement Learning}

Although IRL~\citep{apprentice,maximum} has made significant advances in IL, it encounters some challenges. First, there is an unstable training issue for adversarial learning; improved algorithms have been proposed to overcome this problem. GAIL \citep{gail} is the first study drawing an analogy between IL and generative adversarial networks~\citep{gan}. AIRL~\citep{airl} proposed an AIL method that is robust to changes in dynamics. VAIL~\citep{vail} improves the stability problem by constraining the information flow in the discriminator. EAIRL~\citep{eairl} reduces the overfitting problem using empowerment (the information gain on action entropies). These algorithms have been suggested to improve the stability and scalability of AIL. 

Second, studies on the sample efficiency of expert demonstrations have not been sufficiently conducted. SAILfO~\citep{SAILfO} covers the necessity of studying the sample efficiency of expert demonstrations, and proposes a simple model-based algorithm. f-GAIL~\citep{fgail} showed that finding an appropriate discrepancy measure during training is better than using a predetermined measure to improve sample efficiency. ASAF~\citep{asaf} is an algorithm in which training the discriminator could perform the role of policy and showed that it helps improve sample efficiency. However, these methods require at least five full trajectories to recover the expert policy on continuous control benchmarks such as the MuJoCo physics engine. Unlike those methods, the proposed method successfully imitates the expert's behavior with less than one full trajectory.

In practice, it is difficult to collect perfectly optimal demonstrations because demonstrations are commonly collected by crowdsourcing~\citep{inference} or multiple experts. The collected data from external sources are normally imperfect---a mixture of optimal and non-optimal demonstrations. To address these problems, algorithms for IL from imperfect demonstrations have been proposed~\citep{2iwil,cail}. We demonstrate that combining the proposed method with other algorithms for IL from imperfect demonstrations further improves them, thus, verifying the scalability of the proposed method.

\section{Theoretical Motivation} \label{sec:analysis}



\textcolor{black}{This section aims to provide the theoretical intuition for designing our cost function and feature space aimed at reducing the generalization gap.}

In a general classification learning process, the data and label spaces are denoted by $\mathcal{X}$ and $\mathcal{Y}$, respectively, and $\mathcal{P}(x, y)$ is the joint distribution of the data and label. The primary objective is to learn a classifier $f: \mathcal{X} \rightarrow \mathcal{Y}$ that minimizes the expected value of a loss term $\ell$ over the joint distribution $\mathcal{P}(x, y)$. The classifier $f$ is composed of a feature extractor $h$ and a fully connected layer $\mathbf{W}$. More formally, when $\mathcal{P}(x, y)$ is known, the expected risk can be defined as follows:
\begin{equation}
    \begin{aligned}
      R(f|\mathcal{P}) = \int \ell(\mathbf{W}^{\top}h(x), y)d\mathcal{P}(x,y).
   \label{eq:expected_risk}
   \end{aligned}
\end{equation}
However, in practice, the data distribution is unknown. Empirical risk minimization (ERM) is commonly utilized to address this problem. Then, the minimization can be defined as follows:
\begin{equation}
    \begin{aligned}
      \hat{R}(f|\mathcal{P}) = \frac{1}{M}\sum_{i=1}^{M}\ell(\mathbf{W}^{\top}h(x_i), y_i),
   \label{eq:erm}
   \end{aligned}
\end{equation}
where $M$ is the number of training data points. The generalization performance of a practical classification learning process is highly dependent on the volume of training data. However, in many real-world scenarios, acquiring sufficient training data is a challenging and time-consuming task.


The generalization of ERM has been theoretically justified based on Vapnik-Chervonenkis (VC) theory~\citep{vc}. The upper bound of the expected risk of the classifier $f$ is formulated by 
\begin{equation}
    \begin{aligned}
      R(f|\mathcal{P}) \leqslant \hat{R} (f|\mathcal{P}) + O\left (\left (\frac{|\mathcal{F}|_{\text{VC}}-\text{log}\: \delta }{M}\right )^{\xi} \right ),
   \label{eq:vc}
   \end{aligned}
\end{equation}
where $0 \leq \delta \leq 1$, $0.5 \leq \xi \leq 1$, and $|\mathcal{F}|_{\text{VC}}$ is the finite VC dimension of a classifier. 


Let $\mathcal{H} \subseteq \mathbb{R}^D$, where $D$ represents latent dimension, be the feature space of $h$ and $\mathbf{W} \in \mathbb{R}^{D \times C}$, where $C$ denotes the number of classes, be the last fully connected layer of $f$.


\begin{assumption} \label{assumption1}
     Consider a perfectly trained fully connected layer $\mathbf{W}^*$. We can then split the features as class-relevant or class-irrelevant based on their mutual information with a class label $Y$. Class-relevant features have high mutual information between $h_d(x)$ for $1 \leq d \leq D$ and $Y$, while class-irrelevant features have low mutual information between $h_d(x)$ for $1 \leq d \leq D$ and $Y$.
\end{assumption}



\begin{definition} \label{definition1}
        \citep{revisited} To minimize a change in classification results by variation of class-irrelevant features, $w_{d, i}$ and $w_{d, j}$ for $1 \leq d \leq D$ and $1 \leq i, j \leq C$ should be similar for class-irrelevant features in the empirical risk on original data $\hat{R}(f|\mathcal{P}) = \frac{1}{M}\sum_{i=1}^{M}\ell(\mathbf{W}^{\top} h(x_i), y_i)$.
\end{definition}
Definition 1 from~\citeauthor{revisited} shows that in $\hat{R}(f|\mathcal{P})$, $w_{d, i}$ for all $i$ for class-irrelevant features are not forced to be 0, that is, $f^*$ preserves class-irrelevant features.



Eq.~\ref{eq:vc} highlights three key factors that can improve the generalization performance of a classifier: 1) minimizing the empirical risk $\hat{R}(f|\mathcal{P})$, 2) increasing the size of the training data, and 3) reducing the VC dimension of the classifier. To improve the generalization, we propose an approach that leverages self-supervised learning. By utilizing transformed data as its own supervision, self-supervised learning can learn feature representations that filter out information about nuisance variables. This enables the model to capture more meaningful and generalizable features that can improve the overall performance of the classifier. 

To begin with, we aim to reduce the upper bound of the expected risk by minimizing the discrepancy between the original and transformed samples in the feature space $\mathcal{H}$. \textcolor{black}{Formally, we utilize the mean squared error (MSE), the most widely-used discrepancy loss, as follows:}
\begin{equation}
    \begin{aligned}
    \mathcal{L} = \frac{1}{M \times G}\sum_{i=1}^{M \times G} \left \| h(x_i) - h(x'_i) \right \|_{2}^{2},
   \label{eq:theorymse}
   \end{aligned}
\end{equation}
\textcolor{black}{where $x'$ represents the transformed version of $x$ and $M \times G$ is the number of training data with a finite constant number of augmentations $G$.}

By minimizing Eq.~\ref{eq:theorymse}, we can effectively learn a feature space that does not encode the distortions present in the input samples due to transformations. In other words, the resulting feature representation is invariant to such distortions, leading to improved generalization performance. Given the feature extractor with Eq.~\ref{eq:theorymse} as $h_{\text{MSE}}$ and the corresponding feature space as $\mathcal{H}_{\text{MSE}}$, the classifier $f_{h_{\text{MSE}}}$ consisting of $h_{\text{MSE}}$ holds the following empirical risk equation:
\begin{equation}
    \begin{aligned}
    R(f_{h_{\text{MSE}}}|\mathcal{P}) \leqslant \hat{R} (f_{h_{\text{MSE}}}|\mathcal{P}) + O\left (\left (\frac{|\mathcal{F}_{h_{\text{MSE}}}|_{\text{VC}}-\text{log}\: \delta }{M}\right )^{\xi} \right ),
   \label{eq:vc_aug}
   \end{aligned}
\end{equation}
where $|\mathcal{F}_{h_{\text{MSE}}}|_{\text{VC}}$ is the finite VC dimension of the classifier with the feature space $\mathcal{H}_{\text{MSE}}$ trained with Eq.~\ref{eq:theorymse} simultaneously.


\begin{corollary} \label{corollary1}
    $h_{\text{MSE}}$ is effective in reducing the VC dimension, however, the problem is that the upper bound of $R(f|\mathcal{P})$ we seek to is based on $\hat{R}(f|\mathcal{P})$ instead of $\hat{R}(f_{h_{\text{MSE}}}|\mathcal{P})$. As a result, $f_{h_{\text{MSE}}}$ cannot capture all the properties preserved in $\hat{R}(f|\mathcal{P})$. Additionally, due to the distribution gap between the feature spaces $h$ and $h_{\text{MSE}}$, the optimal function $f_{h_{\text{MSE}}}^*$ of $\hat{R}(f_{h_{\text{MSE}}}|\mathcal{P})$ is not guaranteed to be a minimum of $\hat{R}(f|\mathcal{P})$.
\end{corollary}
\noindent \textit{Proof.} The proof can be found in Appendix~\ref{sec:appendixsuba1}.





The feature extractor $f_h$ typically preserves both class-relevant and class-irrelevant features, whereas the embedding learned by $f_{h_{\text{MSE}}}$ is more heavily regularized due to the intensive transformation applied to the input samples. This regularization can be beneficial for reducing the upper bound of the expected risk of the classifier, but it can also introduce a distribution shift. To mitigate this issue, unsupervised contrastive learning can be utilized.



We assume that MSE and contrastive learning share the feature space. 


\begin{definition} \label{definition2}
        \citep{hypersphere} Consider a perfectly trained $\mathbf{W}^*$ via contrastive learning. The feature space $\mathcal{H}_{\text{MSE}}$ becomes maximally informative.
\end{definition}


\begin{claim} \label{claim1}
    Class-irrelevant yet informative features are preserved via contrastive learning. It reduces the inconsistency between $\hat{R} (f_{h_\text{MSE}}|\mathcal{P})$ and $\hat{R} (f|\mathcal{P})$.
\end{claim}
\noindent \textit{Proof.} The proof with assumptions can be found in Appendix~\ref{sec:appendixsuba2}.


\textcolor{black}{By combining MSE (Eq.~\ref{eq:theorymse}) with contrastive learning in the feature space, we anticipate a notable enhancement in generalization, even when dealing with a modest amount of training data. Our suggested algorithm expands this theoretical foundation rooted in classification to AIL. Importantly, we capture temporally significant features crucial in RL by creating a transformed version based on temporal features for contrastive learning.}

\begin{figure*}[t]
\centering
\includegraphics[width=1.\linewidth]{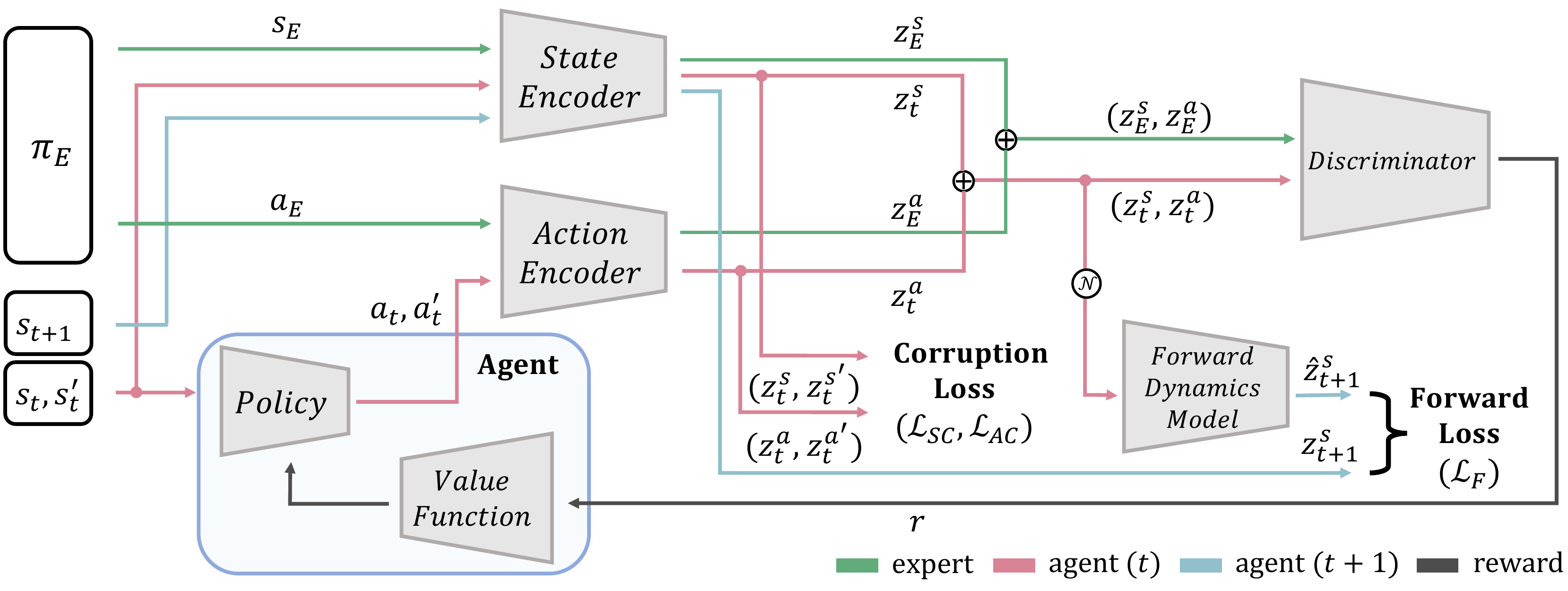}
\caption{\textcolor{black}{Overview of the proposed model. Our proposed model comprises six networks during IRL training. 1) The policy generates actions $a$ based on states $s$ according to a specified policy; 2) The value function evaluates the current policy $\pi_{\theta}$. It is trained using rewards $r$ derived from an estimated cost function (discriminator); 3) The state encoder extracts a feature representation of raw states $s$, 4) The action encoder extracts a feature representation of actions $a$, 5) The forward dynamics model predicts the feature representation of the distorted version of next states $\hat{z}^s_{t+1}$ based on the feature representations of the current state and action, $z^s_t$ and $z^a_t$, along with Gaussian noises $\mathcal{N}$, and 6) The discriminator discriminates agent demonstrations from expert demonstrations. The input is $z^s \oplus z^a$. The discriminator is also referred to as a cost function. More details of each component and loss in the figure are described in Section~\ref{sec:4}.}}
\label{fig:overview}
\vspace{-5mm}
\end{figure*}

\section{Method}\label{sec:4}

The core of RL is an agent and environment. An agent receives a reward from the environment based on the actions determined by a policy. RL learns the optimal policy of a Markov decision process. For IL, the agent learns the optimal policy where a pre-defined reward function does not exist by relying only on given expert demonstrations. We define the IL scenario from a small number of expert demonstrations. The expert demonstrations $\mathcal{D}_E$ are sampled from a state-action density of an expert, $\rho_O$, defined as follows:
\begin{equation}
    \begin{aligned}
      \mathcal{D}_E = \left \{ (s_n, a_n) \right \}_{n=1}^{N_E} \stackrel{i.i.d.}{\sim} \rho_O(s, a),
   \label{eq:noisydataset}
   \end{aligned}
\end{equation}
where $N_E$ is the number of state-action pairs from $\rho_O$. We assume a scenario in which $N_E$ is less than the number of one full trajectory. We denote a state-action pair by $x = (s, a)$ where $x \in X$ and $X = S \times A$.

This study comprises the following six networks, as illustrated in Figure~\ref{fig:overview}.

\begin{itemize}
\item \textbf{A policy $\pi_{\theta}(\cdot)$} parameterized by $\theta$ that generates actions $a$ given states $s$ based on a policy.

\item \textbf{A value function $V(\cdot)$} that evaluates a current policy $\pi_{\theta}$.~$V$ is trained with rewards $r$ from an estimated cost function.

\item \textbf{A state encoder $SE(\cdot)$} that extracts a feature representation of raw states $s$. $z^s = SE(s) \in Z^s$.

\item \textbf{An action encoder $AE(\cdot)$} that extracts a feature representation of actions $a$. $z^a = AE(a) \in Z^a$. 

\item \textbf{A forward dynamics model $F(\cdot)$} that predicts the feature representation of the distorted version of next states $\hat{z}^s_{t+1}$ from the feature representations of the current state and action, $z^s_t$ and $z^a_t$, and Gaussian noises $\mathcal{N}$. $\hat{z}^s_{t+1}$ = $F(z^s_t \oplus z^a_t \oplus \mathcal{N}) \in \hat{Z}_{t+1}$, where $\oplus$ is concatenation.

\item \textbf{A discriminator $D_{\omega}(\cdot)$} parameterized by $\omega$ that discriminates agent demonstrations from expert demonstrations $\mathcal{D}_E.$ The input of $D_{\omega}$ is $z^s \oplus z^a$. $D_{\omega}$ is also called a cost function. In the RL phase, $D_{\omega}$ is the estimated cost function ($r = -\text{log} (D_{\omega}(z^s \oplus z^a)) \in R$).
\end{itemize}

As shown in Algorithm~\ref{alg:algo}, the proposed method comprises three major parts. In Section~\ref{sec:4.1}, we explain how to train the cost function $D_{\omega}$ using expert demonstrations $\mathcal{D}_E$ (GAIL in Algorithm 1). In Section~\ref{sec:4.2}, we describe how to use SSL in a non-image environment (REPR in Algorithm 1). We implement trust region policy optimization~\citep{trpo} to train the agent policy $\pi_{\theta}$ by following the use in~\cite{gail} (TRPO in Algorithm 1). 

\subsection{Generative Adversarial Imitation Learning} \label{sec:4.1}
Our method is based on generative adversarial imitation learning (GAIL)~\citep{gail}. GAIL finds an optimal policy by matching an occupancy measure between expert $E$ and the agent. The optimization equation of GAIL can be derived in the form of the Jensen-Shannon divergence, which is equal to the minimax equation of generative adversarial networks~\citep{gan}. The minimax optimization of GAIL is expressed as follows:
\begin{equation}
    \begin{aligned}
      \underset{\theta}{\text{min}}\: \underset{\omega}{\text{max}} \underset{x \sim \mathcal{D}_{\pi}}{\mathbb{E}} \left [ \text{log}\: D_{\omega}(x) \right ] + \underset{x \sim \mathcal{D}_E}{\mathbb{E}} \left [ \text{log} (1 - D_{\omega}(x)) \right ],
   \label{eq:gail}
   \end{aligned}
\end{equation}
where $\mathcal{D}_{\pi}$ and $\mathcal{D}_E$ are the corresponding demonstrations from an agent policy $\pi_{\theta}$ and expert policy $\pi_{\text{E}}$, respectively. In GAIL, the raw state and action are input to the discriminator. For the proposed discriminator, state and action representations embedded by state and action encoders, $SE(\cdot)$ and $AE(\cdot)$, are input. The discriminator is expressed as follows:
\begin{equation}
    \begin{aligned}
      \underset{\omega}{\text{max}} \underset{x \sim \mathcal{D}_{\pi}}{\mathbb{E}} \left [ \text{log}\: D_{\omega}(z) \right ] + \underset{x \sim \mathcal{D}_E}{\mathbb{E}} \left [ \text{log} (1 - D_{\omega}(z)) \right ],
   \label{eq:ourgail}
   \end{aligned}
\end{equation}
where $z = z^s \oplus z^a$. Here, $z^s$ is a state representation embedded by $SE(s)$, and $z^a$ is an action representation embedded by $AE(a)$.

\begin{algorithm}[t]
\caption{Sample-efficient Adversarial Imitation Learning}
\label{alg:algo}
\begin{algorithmic}[1]
    \State \textbf{input}: Expert demonstrations $\mathcal{D}_E \triangleq \left \{ x_{i} \right \}_{i=1}^{N_E}$, $\#$ of batches B, Training epochs T.

\For{$k \leftarrow $ 1 to T} \Comment{Section~\ref{sec:4}}
    \State Obtain trajectories $\mathcal{D}_k = \left \{ x_{k, i} \right \}_{i=1}^{N}$ using $\pi_{\theta}$
    \State $\pi_{\theta} \leftarrow$ TRPO($\pi_{\theta}, V, D_{\omega}, \mathcal{D}_k$)  \Comment{\cite{gail}}
    \State $SE, AE \leftarrow$ \Call{Repr}{$SE, AE, F, \mathcal{D}_k$}
    \State $D_{\omega} \leftarrow$ \Call{Gail}{$D_{\omega}, SE, AE, \mathcal{D}_k, \mathcal{D}_{E}$}
\EndFor
\Function{Repr}{$SE, AE, F, \mathcal{D}_k$} \Comment{Section~\ref{sec:4.2}}
    \For{$b \leftarrow $ 1 to $B$}
    \State Generate $X'_b$ by Eq.~\ref{eq:corruption} \Comment{Section~\ref{sec:4.2.2} Corruption Method}
    \State Obtain $Z_b$ from $\mathcal{D}_{k, b}$ using ($SE, AE$)
    \State Obtain $Z'_b$ from $X'_b$ using ($SE, AE$)
    \State Update $SE, AE,$ and $F$ by Eq.~\ref{eq:totalloss}
    \EndFor
    \State \Return $SE, AE$  \Comment{Learning state and action representations}
\EndFunction
\Function{Gail}{$D_{\omega}, SE, AE, \mathcal{D}_k, \mathcal{D}_{E}$} \Comment{Section~\ref{sec:4.1}}
    \For{$b \leftarrow $ 1 to $B$}
    \State Obtain $Z_b$ from $\mathcal{D}_{k, b}$ using ($SE, AE$)
    \State Update $SE, AE,$ and $D_{\omega}$ by Eq.~\ref{eq:ourgail}
    \EndFor
    \State \Return $D_{\omega}$ \Comment{Learning cost function}
\EndFunction
\end{algorithmic}
\end{algorithm}

\subsection{State and Action Representations} \label{sec:4.2}

\subsubsection{Modeling forward Dynamics}\label{sec:4.2.1}

Each domain requires different ways of generating self-supervision depending on the properties of the data. For example, BERT~\citep{bert} leverages a training signal by predicting future words from previous words. For RL, the prediction error of a forward dynamics model has been used as an intrinsic reward~\citep{curiosity,selfpredictive}. In tabular data, in contrast to data from an image, it is difficult to create a distorted version of an original input without losing semantic information. Therefore, we posit that maximizing the agreement between the distorted and original ones is more suitable for learning meaningful features than maximizing the agreement between the two distorted versions of input in tabular data. We propose a method that generates a distorted version of the input to learn or discard the desired features and their corresponding loss function for RL.

The proposed method uses the forward dynamics model to predict the distorted version of the next state representation from the given current state and action representations. First, the forward dynamics model is mathematically expressed as follows:
\begin{equation}
    \begin{aligned}
      \hat{z}^s_{t+1} = F(z^s_t \oplus z^a_t \oplus \mathcal{N}),
   \label{eq:foward}
   \end{aligned}
\end{equation}
where $z^s_t = SE(s_t)$, $z^a_t = AE(a_t)$ at a time step $t$, and $\mathcal{N}(0, 1)$ is the Gaussian noise. The output $\hat{z}^s_{t+1}$ represents a distorted version of the observed future representations $z^{t+1}$. The choice of a transformation controls what the representation learns. Thus, we apply a distortion by concatenating Gaussian noise rather than using a corrupted state-action pair because corruption cannot guarantee consistency in information with respect to temporality. 

We use a contrastive loss for training. InfoNCE-based unsupervised contrastive learning (UCL) methods learn a feature representation by maximizing the agreement between differently transformed same input while minimizing that of the rest of the input (negative samples). The learned representation from UCL is invariant in unnecessary details; however, it contains maximal information by maximizing a lower bound on the mutual information between the two views~\citep{hypersphere,contrastivetheory1}. We utilize the InfoNCE loss to obtain as many temporally informative features as possible. The proposed forward dynamics model is trained to maximize the agreement between the distorted and observed next state representations while minimizing that of the rest of the state representations. This is expressed as follows:
\begin{equation}
    \begin{aligned}
      \mathcal{L}_{F} = -{\mathbb{E}} \left [ \text{log} \frac{e^{\text{cs}(\hat{z}^s_{i, t+1}, z^s_{i, t+1})/\tau}}{\zeta} \right ], \: \: \: \: \zeta = \sum_{j=1}^{BS} \mathds{1}_{j \neq i} e^{\text{cs}(\hat{z}^s_{i, t+1}, z^s_{j, t})/\tau} + \sum_{j=1}^{BS} \mathds{1}_{j \neq i} e^{\text{cs}(\hat{z}^s_{i, t+1}, z^s_{j, t+1})/\tau},
   \label{eq:contrastiveIL}
   \end{aligned}
\end{equation}

where $(z^s_t, z^s_{t+1}, \hat{z}^s_{t+1}) \: \sim\:  (Z^s_t, Z^s_{t+1}, \hat{Z}^s_{t+1})$ and cs($u,v$) $=u^{\top}v/||u|| \, ||v||$ ($j$ indexes the state or next state representation in the batch, and $BS$ is the batch size). Following SimCLR~\citep{contrastive:simclr}, we use the other $2(BS - 1)$ representations in the batch as negative samples.

\subsubsection{Corruption Method}\label{sec:4.2.2}

Learning representations that can discard nuisance features is preferable to reduce sample complexity. To help with this, we propose a corruption method that creates a distorted sample showing diverse views that are possible in-distribution. The proposed method swaps the input features of each state-action pair with the input features of the same indices of another state-action pair in a batch. For a batch of state-action pairs sampled from the current policy, $X_b$, we generate a corrupted version $x'_i$ for each state-action pair $x_i$. The corrupted versions of state $s$ and action $a$ are generated, respectively. However, for convenience of rationalization, we explain the method based on the state-action pair $x$.

First, we make a copy of $X_b$ as $X^c_b$ and permute $X^c_b$ by changing the order of each state-action pair in the batch at random, $perm(X^c_b)$. Second, we sample some indices of the state-action pair without replacement, $I \in \left \{0,...,dim(s \oplus a)-1\right \}^q$. $q$ is the number of features to corrupt ($= \left \lfloor c \cdot dim(s \oplus a) \right \rfloor$), where $c$ is a corruption rate ($c = c^s + c^a$). Third, we duplicate $I$ as a shape of $N(X_b) \times q$. Subsequently, we generate corrupted state-action pairs $X'$ of a given batch as follows: 
\begin{equation}
    \begin{aligned}
      & X^c_b[I] = perm(X^c_b)[I], \\
      & X'_b = X^c_b,
   \label{eq:corruption}
   \end{aligned}
\end{equation}
where $X'_b = \left \langle S'_b \times A'_b \right \rangle$. For convenience, we omit subscripts $b$ hereafter. We refer to this method as the swapping corruption method. Empirically, we observed superior performance compared to existing methods on non-image control benchmarks.

Numerous discrepancy measures can quantify the similarity between corrupted and original inputs. For the state, we maximize the similarity between the representation of the distorted and observed versions by minimizing the mean squared error (MSE) as follows:
\begin{equation}
    \begin{aligned}
      \mathcal{L}_{SC} = {\mathbb{E}} \left \| z^s - z^{s'} \right \|_{2}^{2},
   \label{eq:sc}
   \end{aligned}
\end{equation}
where $(z^s, z^{s'}) \: \sim\:  (Z^s, Z^{s'})$ and $Z^s = SE(S)$ and $Z^{s'} = SE(S')$. For the state representation, temporally predictive features should be embedded well to minimize $\mathcal{L}_F$ simultaneously. Thus, we observed the best performance with MSE compared to other indirect discrepancy measures, as shown in Tab~\ref{tab:all_loss}. For the action representation, because the gradients from the $\mathcal{L}_F$ are not sufficient to hinder collapse, we use the Barlow twins loss~\citep{barlow} that is robust against the constant embedding problem. It has been proven through connection with mutual information that the Barlow twins loss also learns the feature representation which is invariant to the distortion of the sample~\citep{barlow}, similar to the MSE. The Barlow twins loss draws the cross-correlation matrix close to the identity matrix. This is expressed as follows:
\begin{equation}
    \begin{aligned}
      \mathcal{L}_{AC} =  \sum_i \left ( 1 - \mathcal{C}_{ii} \right )^2 - \sum_i \sum_{j\neq i} \mathcal{C}_{ij}^{\: \: \: 2}, \: \: \: \: \: \: \: \: \: \:  \mathcal{C}_{ij} = \frac{ Z_{i}^{a} Z_{j}^{a'}}{\sqrt{ (Z_{i}^{a})^2} \sqrt{ (Z_{i}^{a'})^2}}
   \label{eq:ac}
   \end{aligned}
\end{equation}

where $Z^a = AE(A)$ and $Z^{a'} = AE(A')$. $i, j$ index the vector dimensions of the action representations. Therefore, the Barlow twins loss prevents collapse by maximizing the similarity between the representation of the distorted and observed versions of action and reducing entanglement between the components of the representations. Ablations that can give more intuitions about losses are given in Section~\ref{sec:ablation}. Consequently, the total loss for SSL is computed as follows.
\begin{equation}
    \begin{aligned}
      \mathcal{L}_{SS} = \lambda_F \mathcal{L}_{F} + \lambda_S \mathcal{L}_{SC}+ \lambda_A \mathcal{L}_{AC},
   \label{eq:totalloss}
   \end{aligned}
\end{equation}
where $\lambda_F$, $\lambda_S$, and $\lambda_A$ are hyperparameters for each loss. 

 \label{sec:method}
\section{Experiments}

\begin{table*}[t]
\centering
\caption{Final performance using 100 expert state-action pairs on Ant-v2, HalfCheetah-v2, and  Walker2d-v2 of MuJoCo. Best results are in \textbf{bold}. The proposed method outperforms existing IRL methods by a significant margin. It succeeds at imitating the expert's behavior on all three benchmarks using only 100 expert state-action pairs.}
\begin{adjustbox}{width=1.\linewidth}
\begin{tabular}{@{}ccccccccc@{}}
\toprule
               & BC &GAIL      & AIRL        & VAIL        & EAIRL         & SQIL         & ASAF         & Ours \\ \cmidrule(l){1-1} \cmidrule(l){2-9}
Ant         & 932.2$\pm$171.7 & 4198.2$\pm$72.6 & 3922.9$\pm$210.7 & 4216.8$\pm$31.0 & 3137.5$\pm$424.8 & -141.4$\pm$427.6 & 1015.6$\pm$107.0 & \textbf{ 4554.8$\pm$162.6 } \\
HalfCheetah & 1875.2$\pm$1623.3 & 2034.6$\pm$2384.6 & -214.1$\pm$45.2 & -1012.8$\pm$497.1 & 6.6$\pm$15.0 & -238.0$\pm$22.5 & 1187.6$\pm$1935.9 &  \textbf{5416.0$\pm$203.8 }  \\
Walker2d      & 535.5$\pm$134.4 & 3513.4$\pm$172.9 & 909.7$\pm$695.8 & 3466.7$\pm$109.0 & 2084.9$\pm$2499.7 & 283.3$\pm$26.5 & 192.9$\pm$58.5 &  \textbf{3527.6$\pm$131.4 }  \\ \cmidrule(l){1-1} \cmidrule(l){2-9}
  Average & 1114.3$\pm$688.1 & 3248.8$\pm$876.7   & 1539.5$\pm$317.2    & 2223.6$\pm$212.4    & 1743.0$\pm$979.8    & -32.0$\pm$158.9    & 798.7$\pm$700.4   & \textbf{4499.4$\pm$165.9 } \\
\bottomrule
\end{tabular}
\label{tab:final_performance2}
\end{adjustbox}
\vspace{-2mm}
\end{table*}

\begin{table}[t]
\centering
\caption{Final performance when using the proposed corruption method (Swapping) and existing methods ($N_E$ = 100), and variance and predicted local outliers~\citep{lof} of corrupted states. For measuring the outlier factor of corrupted states, we make use of 10 neighbors from observed states.}
\begin{adjustbox}{width=.8\linewidth}
\begin{tabular}{ccccc}
\toprule
Ant & Random & Mean & Each dim & Swapping  \\
\cmidrule(l){1-1} \cmidrule(l){2-5}
 Cumulative rewards &    4263.7$\pm$243.9   & 4482.0$\pm$127.4     & 4459.3$\pm$128.4     & \textbf{4554.8$\pm$162.6}   \\
 \cmidrule(l){1-1} \cmidrule(l){2-5}
Variance  $\uparrow$     & 0.765$\pm$0.05 & 0.756$\pm$0.03 & \textbf{0.843$\pm$0.02} & \textbf{0.843$\pm$0.02} \\
Predicted local outliers ($\%$) $\downarrow$ & 90$\%\pm$8$\%$ & \textbf{6$\%\pm$5$\%$} & 26$\%\pm$8$\%$ & 11$\%\pm$6$\%$  \\
\bottomrule
\end{tabular}
\label{tab:corruption2}
\end{adjustbox}
\vspace{-3mm}
\end{table}


The efficacy of the proposed approach is assessed using MuJoCo~\citep{mujoco} and Atari RAM of OpenAI Gym~\citep{openai}, with each benchmark having fewer than 100 expert state-action pairs. In scenarios with a limited number of perfect or imperfect expert demonstrations, our method surpasses previous Adversarial Imitation Learning (AIL) approaches by a significant margin. We assessed the performance of the proposed method on five continuous control benchmarks simulated by MuJoCo (Ant-v2, HalfCheetah-v2, Hopper-v2, Swimmer-v2, and Walker2d-v2) in four distinct settings: using expert demonstrations of less than one full trajectory with the optimality of 25$\%$, 50$\%$, 75$\%$, or 100$\%$. We tested the sample efficiency of the proposed method in a scenario where optimal demonstration samples of less than one full trajectory are available ($\leq 100$). Expert demonstrations with optimalities of 25$\%$, 50$\%$, and 75$\%$ represent imperfect demonstrations - a mixture of optimal and non-optimal demonstrations. Imperfect demonstrations $\mathcal{D}_I$ are sampled from a noisy state-action density $\rho$, expressed as follows: $\mathcal{D}_I = \left \{ (s_n, a_n) \right \}_{n=1}^{N_I} \stackrel{i.i.d.}{\sim} \rho(s, a)$, where $N_I$ is the number of state-action pairs from $\rho$. The noisy state-action density $\rho$ can be expressed as follows:
\begin{equation}
    \begin{aligned}
      \rho(s, a) & = \psi \rho_O(s, a) + \sum_{i=1}^{n} v_i \rho_i (s, a) \\
      & = \psi \rho_O(s, a) + (1 - \psi) \rho_N (s, a),
   \label{eq:noisydensities}
   \end{aligned}
\end{equation}
where $\rho_O$ is the state-action density of an expert, $\rho_i$ is the state-action density of a single non-expert, and $n$ is the number of non-experts. Furthermore, $\psi$, satisfying $0 < \psi < 1$, is an unknown mixing coefficient of the optimal and non-optimal state-action densities, and $\psi + \sum_{i=1}^{n}v_i = 1$; that is, an optimality of 25$\%$ indicates that $\psi =$ 0.25. More details on the experimental setting, definition, and hyperparameters can be found in Appendix~\ref{sec:appenB}.


\textbf{Optimality of 100$\%$} $\hspace{2mm}$ First, we evaluated the proposed method with a small number of perfect expert demonstrations. We compared the method with seven existing IL methods: BC, GAIL, AIRL, VAIL, EAIRL, SQIL, and ASAF. Table~\ref{tab:final_performance2} shows that the proposed method outperforms other IL methods on all three benchmarks. Particularly, the proposed method succeeded in perfectly imitating the expert policy on HalfCheetah. Notably, GAIL and VAIL show higher performance than the recently proposed ASAF in 100 state-action pairs. For f-GAIL, we conducted experiments using the official GitHub; however, it failed to converge in less than one full trajectory. We surmise that this is because a reasonable number of expert state-action pairs must be guaranteed to automatically learn an appropriate discrepancy measure for the given pairs. 

We also tested our method with varying expert data sizes. As provided in Table~\ref{tab:ablationexpert} in Appendix~\ref{sec:appendixd}, there is a relatively small decrease in the performance up to $N_E$ = 20 on all three benchmarks. However, when $N_E$ = 10, the performance is decreased by a large margin. To make the experimental results stronger, we tested the reliability of the reported average using IQM~\citep{IQM}. We obtained IQM 4555.8 on Ant, IQM 5420.3 on HalfCheetah, and IQM 3527.9 on Walker2d, which are very close to the reported average. 

In addition, we tested the proposed method (without $\mathcal{L}_{AC}$) on two discrete control benchmarks of OpenAI Gym~(\citealp{openai}; BeamRider-ram-v0, and SpaceInvaders-ram-v0). We observed that the cumulative rewards of the proposed method are superior to those of the GAIL. For the results on the two discrete control benchmarks, please refer to Appendix~\ref{sec:appendixf}. Moreover, the average cumulative rewards of the expert policy that we obtained can be found in Appendix~\ref{sec:appendixsubb1}. 




\textbf{Corruption method} $\hspace{2mm}$  \textcolor{black}{We showed the performance of the swapping corruption method by comparing it with the existing two corruption methods and additionally, a mean corruption method, which replaces the features with the empirical marginal distribution’s mean, in cumulative rewards, variance, and local outlier score.} Table~\ref{tab:corruption2} shows that the proposed method shows a higher performance compared with the three corruption methods. We observed that the proposed method generated transformed samples that provide more diverse views compared with random and mean methods and comparably diverse views compared with the method, obtaining each replaced feature from varying state-action pairs (Each dim). We measured its diversity on the corrupted states using variance $\sigma^2: \frac{1}{1000} \sum_{i=1}^{1000}\left (\sum_{j=1}^{dim(s)}(s'_{i,j} - \bar{s}'_{i,j})^2\right )$ where $\bar{s}'$ denotes the average of $\left \{ s'_i \right \}_{i=1}^{1000}$. 

Also, one potential concern about using corruption as a transformation technique is that the corrupted samples are out-of-distribution, resulting in performance degradation. To evaluate this, we computed the local outlier factor~\citep{lof} of the corrupted states. We computed the percentage of corrupted samples that are local outliers with respect to the observed states.~Table~\ref{tab:corruption2} shows that for the random method, the most corrupted states are predicted as outliers, despite the low variance. For the mean method, the corrupted states are mostly realistic; however, the diversity is lower compared with the swapping method, which limits the effectiveness of data augmentation. Replacing each feature with a feature from a different combination has diversity, but it also creates more out-of-distribution data. The proposed swapping method showed high variance and a relatively low percentage of local outliers. \textcolor{black}{As a result, we empirically confirmed that for control tasks, it can be beneficial to create meaningful in-distribution data in the corruption process.}

\begin{figure*}[t]
\centering
\includegraphics[width=.9\linewidth]{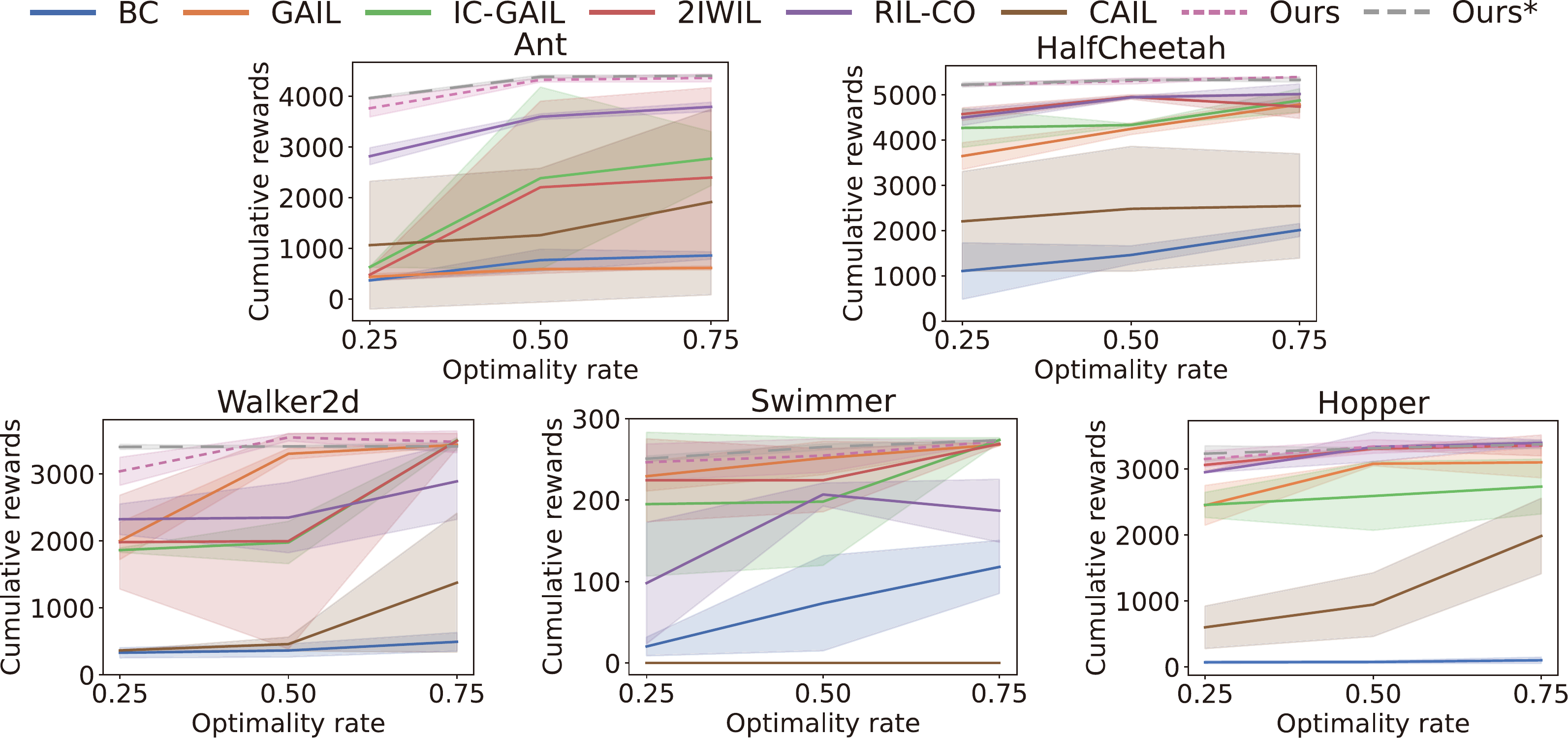}
\caption{Final performance on five continuous control benchmarks with different optimality rates $\psi$. Vertical axes denote cumulative rewards acquired during the last 1000 training iterations. Shaded regions denote standard errors over three runs. Ours* = Ours + MM}
\label{fig:final_performance}
\vspace{-2mm}
\end{figure*}

\begin{table*}[t]
\centering
\caption{Final performance using 25 optimal and 75 non-optimal state-action pairs ($\psi$ = 0.25) to test improvement in sample efficiency.}
\begin{adjustbox}{width=.55\linewidth}
\begin{tabular}{ccccc}
\toprule
 \multicolumn{2}{c}{Ablation}                & \multicolumn{3}{c}{Cumulative Rewards}            \\  \cmidrule(l){1-2} \cmidrule(l){3-5}  
 MM & Ours & Ant & HalfCheetah & Walker2d  \\
 \cmidrule(l){1-2} \cmidrule(l){3-5}
    &        & 478.4$\pm$159.7       & 4573.1$\pm$194.4       &  1981.8$\pm$989.7 \\
\checkmark      &        & 776.3$\pm$1805.1      & 4728.3$\pm$172.8       &  \textbf{3436.0$\pm$15.8} \\
     &    \checkmark     & 3764.6$\pm$241.2       & 5217.4$\pm$29.7      &  3039.6$\pm$295.6 \\
\checkmark      & \checkmark      & \textbf{3966.0$\pm$26.2}        & \textbf{5221.0$\pm$75.1}   &  3403.0$\pm$52.5 \\
 \bottomrule
\end{tabular}
\label{tab:ablation_mm}
\end{adjustbox}
\end{table*}


\textbf{Optimality of 25$\%$, 50$\%$, or 75$\%$} $\hspace{2mm}$ We tested the performance of the proposed method in a more practical scenario with imperfect demonstrations. In this environment, we combined the proposed method with 2IWIL~\citep{2iwil}. 2IWIL showed stable training compared to other IL algorithms for imperfect demonstrations because they predict the confidence of given demonstrations in a pre-stage. Please refer to Appendix~\ref{sec:appendixsubb2} for detailed explanations about 2IWIL, other comparison methods, and the collected imperfect demonstrations. The pseudo-code of the combined algorithm can be found in Appendix~\ref{sec:appendixc}. 

Figure~\ref{fig:final_performance} shows the cumulative rewards on the five continuous control MuJoCo benchmarks with different optimality rates. The proposed method combined with 2IWIL outperforms the other six comparisons by a large margin in all optimality rates. Particularly, the relative improvement over 2IWIL and CAIL~\citep{cail} on Ant is 288$\%$ and 208$\%$ on average, respectively. However, we observed a decrease in the cumulative rewards when the noise rate was 0.75 on Ant and Walker2d benchmarks. We surmise that this was caused by a deficiency in the number of locomotion movements from the optimal policy. The degree of improvement in performance is dependent on a number of given optimal demonstrations to some extent because SSL methods create an auxiliary training signal by leveraging the given data.




\textbf{Manifold mixup} $\hspace{2mm}$ We applied a widely-used sample efficiency technique, manifold mixup (MM)~\citep{mix:manofold_Mixup}, to the combined method. In previous studies~\citep{manifoldmixupselfsupervised,imix}, it is shown that MM in the feature space enriched by SSL is further effective to improve performance. Many mixup-based augmentation studies have been proposed to improve the generalization of image domains~\citep{mix:cutmix,mix:augmix,mix:comix,mix:supermix} and sequential domains~\citep{mix:seqeuntial_text,mix:sequential_sentence}. However, for RL, the application of Mixup is only applicable to image environments~\citep{mix:rl_reg,mix:rl_robotics}. Through this experiment, we would like to compare the efficiency of 1) MM, 2) the proposed method, and 3) using both as a sample efficiency technique. MM increases the diversity of expert demonstrations by interpolating the feature space output of the input data pair. We performed MM on the feature space as follows: $(\bar{z}, \bar{y}) = (\text{Mix}_{\lambda}(z_i, z_j), \text{Mix}_{\lambda}(y_i, y_j))$, where $\text{Mix}_{\lambda}(a, b) = \lambda \cdot a + (1 - \lambda) \cdot b$. Here, ($z_i, z_j$) are the feature representations of ($x_i, x_j$), and ($y_i, y_j$) are the estimated confidence by 2IWIL. Table~\ref{tab:ablation_mm} shows that, when only MM was used, the performance on Walker2d improves; however, the performance on Ant and HalfCheetah does not show a reasonable improvement. This indicates that it is difficult to naturally learn representations that are temporally predictive and robust to diverse distortions from training signals generated by synthetic data from MM. Conversely, when only the proposed method was used, we observed a relatively small increase on Walker2d due to a deficiency in the number of optimal locomotion movements. Consequently, as shown in Figure~\ref{fig:final_performance} and Table~\ref{tab:ablation_mm}, we observed near-optimal performance on all the benchmarks with varying optimalities when both ours and MM were used.


\begin{table*}[t]
\centering
\caption{Ablation studies using 100 expert state-action pairs to test the importance of $\mathcal{L}_F$ and ($\mathcal{L}_{SC}$, $\mathcal{L}_{AC}$). FD = Forward dynamics, and CR = Corruption.}
\begin{adjustbox}{width=.55\linewidth}
\begin{tabular}{ccccc}
\toprule
 \multicolumn{2}{c}{Ablation}                &    \multicolumn{3}{c}{Cumulative Rewards}         \\  \cmidrule(l){1-2} \cmidrule(l){3-5}  
 $\mathcal{L}_F$ & $\mathcal{L}_{SC}$, $\mathcal{L}_{AC}$ & Ant & HalfCheetah & Walker2d  \\
\cmidrule(l){1-2} \cmidrule(l){3-5}  
   &        &  4198.2$\pm$72.5   & 2034.6$\pm$2384.6  & 3513.4$\pm$172.9 \\
\checkmark      &        &    3329.6$\pm$513.0  & 2330.3	$\pm$3741.3   & 3524.5$\pm$17.3 \\
    &  \checkmark    &   2244.0$\pm$208.6   &  591.9$\pm$549.9  & 1028.6$\pm$15.4 \\
\checkmark      & \checkmark      & \textbf{4554.7}$\pm$\textbf{162.5}   & \textbf{5415.9}$\pm$\textbf{203.8}   & \textbf{3527.5}$\pm$\textbf{131.3}   \\
 \bottomrule
\end{tabular}
\label{tab:ablation_forward}
\end{adjustbox}
\end{table*}

\begin{table*}[t]
\centering
\caption{Ablation studies using 50 optimal and 50 non-optimal state-action pairs to test sensitivity to corruption rate of state and action.}
\begin{adjustbox}{width=1.\linewidth}
\begin{tabular}{cccccccc}
\toprule
$c^a$ & \multicolumn{5}{c}{0.2} & 0.5 & <= 0.5  \\ 
\cmidrule(l){1-1} \cmidrule(l){2-6} \cmidrule(l){7-7} \cmidrule(l){8-8}
$c^s$ & 0.1 & 0.2 & 0.3 & 0.4 & <= 0.5 & <= 0.5 & <= 0.5  \\
\cmidrule(l){1-1} \cmidrule(l){2-8}
Ant &   \textbf{4384.71$\pm$49.21}   & 4282.89$\pm$170.53     & 4206.76$\pm$502.76    & 4127.30$\pm$451.87 & 3546.28$\pm$487.71    & 2425.96$\pm$65.69      & 560.64$\pm$326.97  \\
\bottomrule
\end{tabular}
\label{tab:rate}
\end{adjustbox}
\end{table*}

\section{Building Intuitions with Ablations}

We conducted ablations on the proposed factors to provide an intuition of each role. 


\textbf{Importance of both $\mathcal{L}_F$ and ($\mathcal{L}_{SC}$, $\mathcal{L}_{AC}$)}  $\hspace{2mm}$ Table~\ref{tab:ablation_forward} shows that $\mathcal{L}_F$ and ($\mathcal{L}_{SC}$, $\mathcal{L}_{AC}$) are complementary to each other. When only $\mathcal{L}_F$ is added to GAIL, we observed an increase on HalfCheetah and Walker2d, not on Ant, and the gap in increase is small. \citet{makes} demonstrated that it is not always good to learn maximal information using contrastive learning. Rather, it is important to minimize nuisance information as much as possible using a strong transformation to maximize task-specific information. However, a strong transformation without damaging semantic information is impossible in continuous control data. Thus, $\mathcal{L}_F$ mainly plays the role of maximally learning temporally predictive information with a weak, reasonable transformation, and ($\mathcal{L}_{SC}$, $\mathcal{L }_{AC}$) helps to suppress the nuisance factors using the corruption method so as to maximize task-relevant features in the proposed method.

\textbf{Sensitivity to corruption rate} $\hspace{2mm}$ Recently, studies on the importance of the degree of transformation have been proposed in SSL. According to~\citet{makes}, using previously suggested transformation techniques without searching can lead to an increase in learning nuisance information. \citet{understanding} showed that a strong transformation can cause dimensional collapse. They showed that a greedy search for finding appropriate corruption rates is important in applying SSL to RL benchmarks. Table~\ref{tab:rate} shows that the action is very vulnerable to a high corruption rate. For the state, the performance is maintained to some extent below 0.4. Additionally, we confirmed that fixing the corruption rate is better than providing a range.

\textbf{Loss function of $\mathcal{L}_F$} $\hspace{2mm}$ For the discriminator $D_{\omega}$, temporally predictive features are important information to distinguish the imitator (agent) from the expert. However, learning maximal information can lead to learning nuisance information as well. Transformation techniques should be used to suppress this. For $\mathcal{L}_F$, we tested the concatenation of Gaussian noise and the corrupted state as transformation techniques to generate the distorted version. The concatenation of Gaussian noise yields better results. Qualitatively, there was a 7.5$\%$ relative improvement when using the concatenation of Gaussian noise. We surmise that this is because corruption can unavoidably change important input features with respect to temporality. Also, we confirmed that appending noise dimensions rather than no noise concatenation increases the performance (Appendix~\ref{sec:appendixe}). For the loss function of $\mathcal{L}_F$, in addition to the InfoNCE loss of SimCLR, we tested MSE, and MSE with the stop-gradient of BYOL, and the Barlow twins loss. Table~\ref{tab:forward_loss} presents that SimCLR's performance is superior to other methods by a significant margin.


\begin{table}[t]
\begin{minipage}[c]{0.33\textwidth}
\tiny
\centering
\vspace{-1.em}
\caption{Ablation studies us-\\ing 50 optimal and 50 non-\\optimal state-action pairs \\on Ant-v2 to test the role of \\a loss function of $\mathcal{L}_F$.}
\begin{tabular}{cc}
\toprule
 & $\mathcal{L}_F$  \\
\midrule
MSE &    2793.48$\pm$98.59   \\
BYOL &    752.19$\pm$51.08   \\
Barlow twins &    3602.71$\pm$1164.86     \\
SimCLR  &  \textbf{4384.71$\pm$49.21}  \\
\bottomrule
\end{tabular}
\label{tab:forward_loss}
\end{minipage}
\begin{minipage}[c]{0.66\textwidth}
\tiny
\centering
\caption{Ablation studies using 50 optimal and 50 non-optimal state-action pairs on Ant-v2 to test the role of a loss function of $\mathcal{L}_{SC}$ and $\mathcal{L}_{AC}$. BT = Barlow twins.}
\begin{tabular}{@{}c|c|ccccc|c@{}}
\cmidrule(l){3-8}
\multicolumn{2}{l}{} & \multicolumn{6}{c}{$\mathcal{L}_{SC}$} \\ \cmidrule(l){3-8}
\multicolumn{1}{l}{} && MSE & BYOL & SimSiam & BT & VICReg & Avg. \\  \midrule 
\multirow{5}{*}{$\mathcal{L}_{AC}$}
& MSE         & 3658.97	& -627.01 & 1516.06 &	2891.03 &	701.55    &   1628.12  \\
& BYOL        & 2668.32	& 1240.80 & 	1315.78 & 	265.13 & 	902.04    &  1278.41 \\
& SimSiam     & 2717.81	& 4017.71 & 	3854.55 & 	3656.74 &	2943.91 &	3438.14 \\
& BT & $\textbf{4384.71}$	& 2871.37 & 	2590.40 & 	3474.28 & 	4269.20   & \textbf{3517.99}  \\
& VICReg      &  1927.16	& 2363.05 & 	4027.86 & 	3768.11 & 	3458.10   &  3108.86 \\ \midrule
& Avg.     &  \textbf{3071.40} &	1973.18 &	2660.93 &	2811.06 &	2454.96	   &  \\ \bottomrule
\end{tabular}
\label{tab:all_loss}
\end{minipage}
\vspace{-3mm}
\end{table}

\textbf{Loss function of $\mathcal{L}_{SC}$ and $\mathcal{L}_{AC}$} $\hspace{2mm}$ As shown in Table~\ref{tab:all_loss}, we tested various SSL loss functions for both $\mathcal{L}_{SC}$ and $\mathcal{L}_{AC}$. Notably, for $\mathcal{L}_{SC}$, the MSE loss that is exposed to the collapsing problem shows the highest performance on average. This is because $\mathcal{L}_{F}$ cannot be minimized if the state representation is only the same constant. Rather, the MSE loss that can flow the gradients to both the input pair shows a higher performance than the loss functions using a stop-gradient or different discrepancy measures. For $\mathcal{L}_{AC}$, the Barlow twins and SimSiam losses showed the first- and second-best performance on average, respectively. Because the role of the state is greater than that of the action when predicting the next state, the action representation $z^a$ is not completely free from the collapsing problem. Therefore, unlike the trend of $\mathcal{L}_{SC}$, clearly, the loss functions that have a certain technique to prevent it showed stable performance. 
 \label{sec:ablation}
\section{Conclusion}

Sample efficiency of expert demonstrations is desirable in imitation learning because obtaining a large number of expert demonstrations is often costly. Motivated by successes in self-supervised learning, we proposed a sample-efficient imitation learning method that promotes learning feature representations that are temporally predictive and robust against diverse distortions for continuous control. We evaluated our proposed method in various control tasks with limited expert demonstration settings and showed superior performance compared to existing methods. We analyzed the efficiency of the proposed method through both theoretical motivation and extensive experiments on continuous and discrete-control benchmarks. 

Despite the excellent performance with limited settings, the proposed method has some limitations. There is an increase in model complexity and additional computational cost during training since three additional networks and self-supervised losses are added. 





\section{Acknowledgement}

\textcolor{black}{This work was supported by the National Research Foundation of Korea (NRF) grant funded by the Korea government (MSIT) (2022R1A3B1077720), Institute of Information \& communications Technology Planning \& Evaluation (IITP) grant funded by the Korea government (MSIT) [2021-0-01343: Artificial Intelligence Graduate School Program (Seoul National University) and 2022-0-00959] and the BK21 FOUR program of the Education and Research Program for Future ICT Pioneers, Seoul National University in 2023.}


\bibliography{reference}

\newpage

\appendix

\section{Proof for Section~\ref{sec:analysis}} \label{sec:appendixa}

\subsection{Proof of Corollary 1} \label{sec:appendixsuba1}

Using the first-order Taylor approximation, we can expand $\hat{R}(f_{h_\text{MSE}}|\mathcal{P})$. 


\begin{theorem} \label{theorem1}
    Based on \cite{kernel}, picking up $\mathbf{z}_{0} = \underset{\mathbf{z}_i \sim h_{\text{MSE}}(X_i)}{\mathbb{E}}[\mathbf{z}_i]$, the first-order approximation of $\hat{R}(f_{h_\text{MSE}}|\mathcal{P})$ is $\frac{1}{M}\sum_{i=1}^{M}\ell \left ( \mathbf{W}^\top \underset{\mathbf{z}_i \sim h_{\text{MSE}}(X_i)}{\mathbb{E}} [\mathbf{z}_i], y_i \right )$. Then, the first-order effect is that it averages the distorted features that are not necessarily present in the original dataset $\mathcal{P}$. 
\end{theorem}

\noindent \textit{Proof.} $\;\;\;$  For proof, we define a kernel classifier $K$ to map a feature space. The kernel is always positive definite, and its space is expressed through a dot product. The kernel defines various functions, and its space of such functions is called the reproducing kernel Hilbert space (RKHS). 

When we define that $T(x)$ is the probability density that $x$ inherently contains a distortion that is not necessarily present in the original dataset and a kernel $K'$ represents a new feature space via Eq.~\ref{eq:theorymse}, we can have the following proof.

\begin{equation*}
    \begin{aligned}
      K' (x, \tilde{x}) & = \left \langle \mathbb{E}_{x \sim T(x)} [h_{\text{MSE}}(x)],  \mathbb{E}_{\tilde{x} \sim T(\tilde{x})} [h_{\text{MSE}}(\tilde{x})] \right \rangle \\ 
      & = \int_{x \in \mathbb{R}^n} \int_{\tilde{x} \in \mathbb{R}^n} \left \langle h_{\text{MSE}}(x), h_{\text{MSE}}(\tilde{x}) \right \rangle T(x) T(\tilde{x})d\tilde{x} dx \\
      & = \int_{x \in \mathbb{R}^n} \int_{\tilde{x} \in \mathbb{R}^n} K(x, \tilde{x}) T(x) T(\tilde{x})d\tilde{x} dx \\
      & = (TKT^\top ) (x, \tilde{x}),
   \label{eq:risk}
   \end{aligned}
\end{equation*}
where $x$ and $\tilde{x}$ are two inputs. It shows that the proposed approach can be interpreted as a linear classifier $\mathbf{W}$ on the new kernel $K'$. This proof is similar to how a support measure machine (SMM) trains the model with a mean function in the RKHS~\citep{smm}. SMM is proved to be invariant on the distorted features that are not necessarily present in the original dataset $\mathcal{P}$. In the newly learned feature space through Eq.~\ref{eq:theorymse}, the distorted features, which are inherently contained in the original dataset, become non-informative. As a result, the sample complexity required to learn the classifier $f_{h_{\text{MSE}}}$ is reduced. More formally, a support vector machine (SVM)~\citep{svm} is a special case of SMM. SMM is effective at seeking a halfspace that separates a training set by a large margin. In other words, our proposed approach similar to SMM is also effective in inducing the low VC dimension.

\begin{appendixcorollary} \label{appendixcorollary1}
    $h_{\text{MSE}}$ is effective in reducing the VC dimension, however, the major problem is that the upper bound of $R(f|\mathcal{P})$ we seek to is based on $\hat{R}(f|\mathcal{P})$ instead of $\hat{R}(f_{h_{\text{MSE}}}|\mathcal{P})$. As a result, $f_{h_{\text{MSE}}}$ cannot capture all the properties preserved in $\hat{R}(f|\mathcal{P})$. Additionally, due to the distribution gap between the feature spaces $h$ and $h_{\text{MSE}}$, the optimal function $f_{h_{\text{MSE}}}^*$ of $\hat{R}(f_{h_{\text{MSE}}}|\mathcal{P})$ is not guaranteed to be a minimum of $\hat{R}(f|\mathcal{P})$.
\end{appendixcorollary}

\cite{revisited} showed that the optimal function $f^*$ of $\hat{R}(f|\mathcal{P})$ leverages both class-relevant and -irrelevant features ($w_{d, i} \approx w_{d, j}$ for $1 \leq d \leq D$ and $1 \leq i, j \leq C$). However, when finding  $f_{h_\text{MSE}}^*$, distorted, commonly class-irrelevant, features are restrained to be encoded through Eq.~\ref{eq:theorymse}. Thus, the optimal function $f_{h_\text{MSE}}^*$ of $\hat{R}(f_{h_\text{MSE}}|\mathcal{P})$ is not guaranteed to be a minimum of $\hat{R}(f|\mathcal{P})$ due to a distribution shift.

\subsection{Proof of Claim 1}\label{sec:appendixsuba2}

Suppose that the MSE (Eq.~\ref{eq:theorymse}) and contrastive learning share its feature space $\mathcal{H}$. 


\begin{appendixdefinition} \label{appendixdefinition1}
        \citep{hypersphere} Consider a perfectly trained $h_{\text{MSE}}^*$ with unsupervised contrastive learning. A positive pair has similar features on $\mathcal{H}_{\text{MSE}}$ (Alignment), and the feature space $\mathcal{H}_{\text{MSE}}$ becomes maximally informative (Uniformity).
\end{appendixdefinition}


\begin{appendixdefinition} \label{appendixdefinition2}
     \citep{contrastivetheory1} Contrastive loss is defined using a critic function that approximates density ratios $(\frac{p(x|y)}{p(x)} = \frac{p(y|x)}{p(y)})$ of two random variables $X$ and $Y$. It is proved that contrastive loss is a lower bound on mutual information (MI), $I(X; Y)$. By minimizing the loss, the lower bound on MI is maximized.
\end{appendixdefinition}

Unsupervised contrastive learning can be interpreted as an InfoMax principle, as shown in Definitions 1 and 2. In our terms, minimizing contrastive loss is maximizing the lower bound on $I(h_{\text{MSE}}(x); h_{\text{MSE}}(x'))$ for a positive pair $(x, x') \sim \mathcal{P}_{\text{pos}}$. MI between $h_{\text{MSE}}(x)$ and $h_{\text{MSE}}(x')$ can be re-written as follows:
\begin{equation*}
    \begin{aligned}
      I(h_{\text{MSE}}(x); h_{\text{MSE}}(x')) = H(h_{\text{MSE}}(x)) - H(h_{\text{MSE}}(x) | h_{\text{MSE}}(x')).
   \label{eq:risk}
   \end{aligned}
\end{equation*}
The first term of the right-hand side (RHS) shows that uniformity in Definition 1 is mathematically favored by entropy $H(h_{\text{MSE}}(x))$. The second term of RHS shows that alignment in Definition 1 is mathematically favored by conditional entropy $H(h_{\text{MSE}}(x) | h_{\text{MSE}}(x'))$. As a result, to maximally increase the closeness between all positive pairs in the feature space, all informative features should be semantically embedded and utilized. That is, mapping more informative features induces that more similar inputs have more similar features in $\mathcal{H}$. Through that, informative yet class-irrelevant features are also preserved for contrastive learning. 


\begin{theorem} \label{theorem3}
    Based on \cite{kernel}, with $\psi = \underset{z \sim h_{\text{MSE}}(X)} {\mathbb{E}}[z]$ and $\text{Var} = \mathbb{E} \left [ (X - \mathbb{E}[X])^2 \right ]$, the second-order approximation of $\hat{R}(f_{h_\text{MSE}}|\mathcal{P})$ is the first-order approximation plus variance regularization of weight vectors $w_d$ for $1 \leq d \leq D$.  
\end{theorem}
\noindent \textit{Proof.} $\;$ For cross entropy loss with softmax, $\ell''$ is independent of $y$ and positive semi-definite. We can obtain a more exact expression for $\hat{R}(f_{h_\text{MSE}}|\mathcal{P})$ by the second-order approximation of the Taylor expansion. 

\begin{equation*}
    \small
    \begin{aligned}
      \hat{R}(f_{h_\text{MSE}}|\mathcal{P}) & = \frac{1}{M}\sum_{i=1}^{M}\ell \left ( \mathbf{W}^\top \psi_i, y_i \right ) + \frac{1}{2M} \sum_{i=1}^{M} \underset{z_i \sim h_{\text{MSE}} (X_i) }{\mathbb{E}} \left [ \left ( \mathbf{W}^\top (z_i - \psi_i) \right )^2 \ell'' (\mathbf{W}^\top z_i, y_i) \right ] \\
      & = \frac{1}{M}\sum_{i=1}^{M}\ell \left ( \mathbf{W}^\top \psi_i, y_i \right ) + \\ 
      & \;\;\;\;\;\;\;\;\;\;\;\;\;\;\;\;\;\;\; \mathbf{W}^\top \left ( \frac{1}{2M} \sum_{i=1}^{M} \underset{z_i \sim h_{\text{MSE}} (X_i)}{\mathbb{E}} \left [ \Delta_i \Delta_i^\top \ell'' (\mathbf{W}^\top z_i, y_i) \right ]  \right ) \mathbf{W} \;\;\;\; \left ( \because \Delta_i = z_i - \psi_i \right ) \\
      & = \frac{1}{M}\sum_{i=1}^{M}\ell \left ( \mathbf{W}^\top \psi_i, y_i \right ) + \\
      & \;\;\;\;\;\;\;\;\;\;\;\;\;\;\;\;\;\; \frac{1}{2M} \sum_{i=1}^{M} \mathbf{W}^\top \underset{z_i \sim h_{\text{MSE}} (X_i)}{\mathbb{E}} \left [ \Delta_i \Delta_i^\top \right ]  \ell'' (\mathbf{W}^\top \psi_i) \mathbf{W} \;\;\; (\because \ell'' \; \text{is independent of} \; y) 
   \label{eq:risk}
   \end{aligned}
\end{equation*}

The second-order approximation of $\hat{R}(f_{h_\text{MSE}}|\mathcal{P})$ can be interpreted as the first-order approximation plus variance regularization of weight vectors $w_d$ for $1 \leq d \leq D$ because $\Delta \Delta^\top$ is equal to the variance of $z$. It represents that the weights of features having large variances are forced to 0 restricted by regularization. We do not utilize data augmentation in the classification loss because it further induces the generalization gap. When the feature space is learned through only Eq.~\ref{eq:theorymse} and the classification loss, most class-irrelevant features cannot be encoded in the feature manifold. We tackle this problem by mapping class-irrelevant yet informative features learned from contrastive learning to the feature manifold. For contrastive learning, the positive pair, which is created by two differently augmented same images, should have very similar feature representation. It means informative features mapped through this loss are robust to the variance via augmentation.





\begin{appendixclaim} \label{appendixclaim1}
    Class-irrelevant yet informative features are preserved via unsupervised contrastive learning. It reduces the inconsistency between $\hat{R} (f_{h_\text{MSE}}|\mathcal{P})$ and $\hat{R} (f|\mathcal{P})$.
\end{appendixclaim}

Class-irrelevant yet informative features are preserved via unsupervised contrastive learning. In other words, the feature manifold is enriched by unsupervised contrastive learning, that is, both class-relevant and class-irrelevant yet informative features can be leveraged when finding the optimal function $f_{h_\text{MSE}}^*$ of $\hat{R}(f_{h_\text{MSE}}|\mathcal{P})$. As a result, the inconsistency between $\hat{R} (f_{h_\text{MSE}}|\mathcal{P})$ and $\hat{R} (f|\mathcal{P})$ can be reduced.





\section{Additional Implementation and Experimental Details} \label{sec:appenB}

We use 5 continuous control benchmarks on Mujoco~\citep{mujoco} (Ant-v2, Hopper-v2, HalfCheetah-v2, Swimmer-v2, and Walker2d-v2), and 2 discrete control benchmarks on Atari RAM (BeamRider-ram-v0, and SpaceInvaders-ram-v0) of OpenAI Gym \citep{openai}.

Overall, we reported the mean and standard error of the performance over 3 trials. For experimental settings, we used GTX 1080 Ti for GPUs, Intel i7-6850K for CPUs, and Ubuntu 18.04 for OS. The usage of GPU memory is approximately 3000MB for training, and training time for tested benchmarks was approximately 20 hours. Our code is based on Pytorch~\citep{pytorch} and python libraries.

We make use of the same neural net architecture and hyperparameters for all benchmarks. For the policy network, value network, discriminator, and state encoder, we use 3 hidden layers with size 100 and Tanh as activation functions. For the action encoder, we use 6 1D convolutional layers with sizes (64, 64, 64, 128, 256, and 256), 1 hidden layer with size 8 as the output, and LeakyReLU as activation functions. For the forward dynamics model, we use 1 hidden layer with size 114 and ReLU as activation functions.

For hyperparameters in all runs, the total epoch for Swimmer, and Hopper is 3,000, for BeamRider, SpaceInvaders, HalfCheetah, and Walker2d is 5,000, and for Ant is 8,000. Please refer to Table~\ref{tab:hyp} for other hyperparameters. We set $\lambda_F=1$, $\lambda_S=100$, and $\lambda_A=1$ for matching loss scale.

\begin{table}[h]
\centering
\caption{Base hyperparameters used for all benchmarks.}
\begin{adjustbox}{width=0.6\linewidth}
\begin{tabular}{@{}cc@{}}
\toprule
Hyperparameters                                       & Value          \\ \midrule
$\gamma$                                                 & 0.995          \\
Generalized advantage estimation                      & 0.97           \\
$N$                                        & 5,000          \\
Learning rate (all networks except for value network) & 1e-3           \\
Learning rate (value network)                         & 3e-4           \\
Batch size (RERP)                              & 256            \\
Batch size (TRPO)                                    & 128            \\
Batch size (GAIL)                                     & 5,000          \\
Optimizer (all networks)                              & Adam           \\
$\tau$                                                & 0.1            \\ 
$\lambda_F$                                                & 1            \\ 
$\lambda_S$                                                & 100            \\
$\lambda_A$                                                & 1            \\\bottomrule
\end{tabular}
\label{tab:hyp}
\end{adjustbox}
\end{table}

For the random corruption method~\citep{neurips2020vime}, we sampled an imputing value from $\mathcal{N}(0, 1)$ by considering the range of state values. The mean value used for imputation~\citep{iclr2022scarf} is the mean vector of each $\mathcal{D}_k$ where $k$ is the training iteration. We imputed with a mean value of the corresponding dimension of the calculated mean vector.

\subsection{Optimality of 100$\%$ (Expert Demonstrations)} \label{sec:appendixsubb1}

To train experts, we used a reinforcement learning algorithm, proximal policy optimization \citep{ppo}, uploaded in the official Github by the authors of CAIL \citep{cail}. We selected the converged policy as the expert policy. The performance of the utilized expert policy and other specifications related to the experiments on the main manuscript are given in Table~\ref{tab:spec_continuous}. 


\begin{table}[h]
\centering
\caption{Specification and the number of used demonstrations of each continuous control benchmark in the scenario of perfect expert demonstrations.}
\label{tab:spec_continuous}
\begin{adjustbox}{width=.6\linewidth}
\begin{tabular}{@{}c|cc|c|c@{}}
\toprule
Benchmarks     & dim($S$) & dim($A$) & $N_E$ & Expert's Performance \\ \midrule
HalfCheetah-v2 & $\mathbb{R}^{17}$    & $\mathbb{R}^{6}$    & 100 & 5455.49$\pm$74.26  \\
Walker-v2      & $\mathbb{R}^{17}$    & $\mathbb{R}^{6}$   & 100 & 3685.27$\pm$57.99 \\
Ant-v2         & $\mathbb{R}^{111}$   & $\mathbb{R}^{8}$    & 100 & 4787.23$\pm$115.72 \\
\bottomrule
\end{tabular}
\end{adjustbox}
\end{table}


\subsection{Optimality of 25$\%$, 50$\%$, or 75$\%$ (A Mixture of Optimal and Non-optimal Demonstrations)} \label{sec:appendixsubb2}


We also tested the effectiveness of the proposed method with imperfect demonstrations. We combined the proposed method with the existing method for imperfect demonstrations and checked improvement in performance. We compared our combined method against the following baselines: BC~\citep{alvinn}, GAIL~\citep{gail}, IC-GAIL~\citep{2iwil}, 2IWIL~\citep{2iwil}, RIL-CO~\citep{rgail}, and CAIL~\citep{cail}. 

IL methods, especially variants of BC, require a large volume of expert demonstration data for training~\citep{reduction}. These methods struggle with a generalization problem when using a small number of demonstrations. Empirically, we observed that BC almost fails to converge on Ant, Walker2d, and Hopper with a small number of demonstrations. For RIL-CO, they measure and optimize a classification risk with the symmetric loss. Basically, they only assumed a scenario that the majority of demonstrations are obtained using an optimal policy. IC-GAIL and 2IWIL were proposed in the same paper. Both methods are confidence-based. In the case of 2IWIL, the confidence of each state-action pair is estimated using the classification risk before training, and in the case of IC-GAIL, they implicitly utilize the confidence score in a way of matching the occupancy measure of the imitator with the expert. CAIL is the state-of-the-art work in IL algorithms for imperfect demonstrations. They jointly learn the confidence score and policy using an outer loss. Because they update two factors jointly, the training is unstable. 
Experimentally, there was no benchmark that CAIL, which is most recently suggested, is superior to 2IWIL or RIL-CO with a small number of imperfect demonstrations. We surmise that this is because CAIL is the method using a full trajectory rather than each state-action pair when estimating the confidence score of each pair. Consequently, we decided to combine our method with 2IWIL instead of CAIL.
\begin{table}[h]
\centering
\caption{Specification and the number of used demonstrations of each continuous control benchmark in the scenario of imperfect expert demonstrations.}
\label{tab:spec_continuous2}
\begin{adjustbox}{width=1.\linewidth}
\begin{tabular}{@{}c|cc|c|ccccc@{}}
\toprule
Benchmarks     & dim($S$) & dim($A$) & $N_E$ & Suboptimal 1 & Suboptimal 2 & Suboptimal 3 & Suboptimal 4 & Expert's Performance \\ \midrule
HalfCheetah-v2 & $\mathbb{R}^{17}$    & $\mathbb{R}^{6}$    & 100 & 1051.91$\pm$50.17 & 2280.87$\pm$651.92 & 3533.07$\pm$79.47 & 4682.89$\pm$54.33  & 5455.49$\pm$74.26  \\
Walker-v2      & $\mathbb{R}^{17}$    & $\mathbb{R}^{6}$   & 100 & 691.14$\pm$96.12  & 1617.02$\pm$721.00  & 2579.41$\pm$512.19 & 2819.63$\pm$609.49 & 3685.27$\pm$57.99 \\
Ant-v2         & $\mathbb{R}^{111}$   & $\mathbb{R}^{8}$    & 100 & 789.13$\pm$170.45 & 2115.17$\pm$328.20 & 2947.49$\pm$191.72 & 3739.91$\pm$96.56 & 4787.23$\pm$115.72 \\
Swimmer-v2     & $\mathbb{R}^{8}$     & $\mathbb{R}^{2}$    & 20 & 65.56$\pm$18.93 & 148.20$\pm$8.09 & 181.27$\pm$4.23 & 228.29$\pm$5.95 & 280.5$\pm$1.24 \\
Hopper-v2      & $\mathbb{R}^{11}$    & $\mathbb{R}^{3}$    & 20 & 1262.34$\pm$296.32 & 1774.65$\pm$462.52 & 2185.33$\pm$996.92 & 2802.18$\pm$489.85  & 3531.03$\pm$23.51 \\ \bottomrule
\end{tabular}
\end{adjustbox}
\end{table}

For the experiments on the main manuscript, we collected a mixture of optimal and non-optimal demonstrations with different optimalities. For collecting the imperfect demonstrations, we used the official Github by the authors of CAIL. Following CAIL, We selected four intermediate policies as sub-optimal policies and the converged policy as the optimal policy. The performance of sub-optimal and optimal policies and other specifications for the benchmarks are given in Table~\ref{tab:spec_continuous2}. 
Empirically, as shown in Table~\ref{tab:spec_continuous2}, because the dimension of the action space of Swimmer-v2 and Hopper-v2 is very small almost like discrete control, we did not apply $\mathcal{L}_{AC}$.

The pseudo-code of the combined methods is given in Algorithms~\ref{alg:algo1} and~\ref{alg:algo2}. To run the combined methods, we need additional hyperparameters and their values are summarized in Table~\ref{tab:hyp2}.

\begin{table}[h]
\centering
\caption{Additional hyperparameters for the combined methods.}
\begin{adjustbox}{width=0.4\linewidth}
\begin{tabular}{@{}cc@{}}
\toprule
Hyperparameters                                       & Value          \\ \midrule
$\alpha$ for Mixup        & 4.0            \\
Threshold for GMM            & 0.5            \\
The number of non-experts             & 4            \\
Ratio of labeled demonstrations             & 0.4            \\
\bottomrule
\end{tabular}
\label{tab:hyp2}
\end{adjustbox}
\end{table}

\section{Pseudo-code of Combined Method} \label{sec:appendixc}

\begin{algorithm}[t]
\caption{Pseudo-code of Ours + 2IWIL~\citep{2iwil}}
\label{alg:algo1}
\begin{algorithmic}[1]
    \State \textbf{input}: Imperfect expert demonstrations $\mathcal{D}_I = \left \{\mathcal{D}_L \cup \mathcal{D}_U \right \}$, Labeled demonstrations $\mathcal{D}_L \triangleq \left \{ (x_{l, i}, y_{l, i}) \right \}_{i=1}^{N_L}$, Unlabeled demonstrations $\mathcal{D}_U \triangleq \left \{ x_{u, i} \right \}_{i=1}^{N_U}$, $\#$ of batches B, Training epochs T.
    
    \State Train a probabilistic classifier by minimizing Eq.~\ref{eq:risk}
    \State Predict confidence scores $\left \{ \hat{y}_{u,i} \right \}_{i=1}^{N_u}$ for $\left \{ x_{u, i} \right \}_{i=1}^{N_u}$
\For{$k \leftarrow $ 1 to T}
    \State Obtain trajectories $\mathcal{D}_k = \left \{ x_{k, i} \right \}_{i=1}^{N}$ using $\pi_{\theta}$
    \State $\pi_{\theta} \leftarrow$ TRPO($\pi_{\theta}, V, D_{\omega}, \mathcal{D}_k$)
    \State $SE, AE \leftarrow$ \Call{Repr}{$SE, AE, F, \mathcal{D}_k$}
    \State $D_{\omega} \leftarrow$ \Call{Gail}{$D_{\omega}, SE, AE, \mathcal{D}_k, \mathcal{D}_{I}$}
\EndFor
\Function{Repr}{$SE, AE, F, \mathcal{D}_k$}
    \For{$b \leftarrow $ 1 to $B$}
    \State Generate $X'_b$ by Eq.~\ref{eq:ourgail}
    \State Obtain $Z_b$ from $\mathcal{D}_{k, b}$ using ($SE, AE$)
    \State Obtain $Z'_b$ from $X'_b$ using ($SE, AE$)
    \State Update $SE, AE,$ and $F$ by Eq.~\ref{eq:totalloss}
    \EndFor
    \State \Return $SE, AE$
\EndFunction
\Function{Gail}{$D_{\omega}, SE, AE, \mathcal{D}_k, \mathcal{D}_{I}$}
    \For{$b \leftarrow $ 1 to $B$}
    \State Obtain $Z_b$ from $\mathcal{D}_{k, b}$ using ($SE, AE$)
    \State Update $SE, AE,$ and $D_{\omega}$ by Eq.~\ref{eq:2iwil_gail}
    \EndFor
    \State \Return $D_{\omega}$
\EndFunction
\end{algorithmic}
\end{algorithm}

\begin{algorithm}[t]
\caption{Pseudo-code of Ours + 2IWIL + Manifold Mixup~\citep{manifoldmixup}}
\label{alg:algo2}
\begin{algorithmic}[1]
    \State \textbf{input}: Imperfect expert demonstrations $\mathcal{D}_I = \left \{\mathcal{D}_L \cup \mathcal{D}_U \right \}$, Labeled demonstrations $\mathcal{D}_L \triangleq \left \{ (x_{l, i}, y_{l, i}) \right \}_{i=1}^{N_L}$, Unlabeled demonstrations $\mathcal{D}_U \triangleq \left \{ x_{u, i} \right \}_{i=1}^{N_U}$, $\#$ of batches B, Training epochs T.
    
    \State Train a probabilistic classifier by minimizing Eq.~\ref{eq:risk}
    \State Predict confidence scores $\left \{ \hat{y}_{u,i} \right \}_{i=1}^{N_U}$ for $\left \{ x_{u, i} \right \}_{i=1}^{N_U}$
    \State $\tilde{\mathcal{D}}_{O}, \tilde{\mathcal{D}}_{N} \leftarrow \mathit{GMM}(\mathcal{D}_L, (\mathcal{D}_U, \left \{ \hat{y}_{u,i} \right \}_{i=1}^{N_U}))$ 
\For{$k \leftarrow $ 1 to T}
    \State Obtain trajectories $\mathcal{D}_k = \left \{ x_{k, i} \right \}_{i=1}^{N}$ using $\pi_{\theta}$
    \State $\pi_{\theta} \leftarrow$ TRPO($\pi_{\theta}, V, D_{\omega}, \mathcal{D}_k$)
    \State $SE, AE \leftarrow$ \Call{Repr}{$SE, AE, F, \mathcal{D}_k$}
    \State $D_{\omega} \leftarrow$ \Call{Gail}{$D_{\omega}, SE, AE, \mathcal{D}_k, \tilde{\mathcal{D}}_{O}, \tilde{\mathcal{D}}_{N}$}
\EndFor
\Function{Repr}{$SE, AE, F, \mathcal{D}_k$}
    \For{$b \leftarrow $ 1 to $B$}
    \State Generate $X'_b$ by Eq.~\ref{eq:ourgail}
    \State Obtain $Z_b$ from $\mathcal{D}_{k, b}$ using ($SE, AE$)
    \State Obtain $Z'_b$ from $X'_b$ using ($SE, AE$)
    \State Update $SE, AE,$ and $F$ by Eq.~\ref{eq:totalloss}
    \EndFor
    \State \Return $SE, AE$
\EndFunction
\Function{Gail}{$D_{\omega}, SE, AE, \mathcal{D}_k, \tilde{\mathcal{D}}_{O}, \tilde{\mathcal{D}}_{N}$}
    \For{$b \leftarrow $ 1 to $B$}
    \State Obtain $Z_b$ from $\mathcal{D}_{k, b}$ using ($SE, AE$)
    \State Obtain ($Z_o$, $Z_n$) from ($\tilde{\mathcal{D}}_{O}$, $\tilde{\mathcal{D}}_{N}$) using ($SE, AE$)
    \State Compute $(\bar{Z}, \bar{Y})$ by Eq.~\ref{eq:ourmixup}
    \State Update $SE, AE,$ and $D_{\omega}$ by Eq.~\ref{eq:weightedgail}
    \EndFor
    \State \Return $D_{\omega}$
\EndFunction
\end{algorithmic}
\end{algorithm}

\subsection{Combined Case $\#$1: Ours + 2IWIL}

Because labeling all state-action pairs from $\rho_O$ or $\rho_N$ can be expensive, following previous IL works for imperfect demonstrations, we assumed that only a few demonstrations are labeled. Then, the imperfect demonstrations are split into two demonstrations: a set of labeled demonstrations $\mathcal{D}_L = \left \{ (x_{l, i}, y_{l, i}) \right \}_{i=1}^{N_L}$ and a set of unlabeled demonstrations $\mathcal{D}_U = \left \{ x_{u, i} \right \}_{i=1}^{N_U}$, where $N_L$ and $N_U$ are the number of labeled demonstrations and unlabeled demonstrations, respectively. 

As shown in Algorithm~\ref{alg:algo1}, before training, 2IWIL~\citep{2iwil} estimates pseudo labels $\hat{y}_{u, i}$ of $\mathcal{D}_U$ to utilize the set of unlabeled demonstrations $\mathcal{D}_U$ with $\mathcal{D}_L$ using the classification risk proposed in their work as follows:
\begin{equation}
    \begin{aligned}
      R_{\ell(g)} = \mathbb{E}_{x,\: y\: \sim\:  \mathcal{D}_l}\left [ y(\ell(g(x)) - \ell(-g(x))) + (1 - \beta ) \ell(-g(x)) \right ] + \mathbb{E}_{x\: \sim\: \mathcal{D}_u} \left [ \beta \ell(-g(x)) \right ],
   \label{eq:risk}
   \end{aligned}
\end{equation}
where $\beta = \frac{N_U}{N_L+N_U}$, $g(\cdot)$ is a neural network classifier, and $\ell$ is a strictly proper composite loss. Then, the predicted confidence score $\hat{y}_u$ represents the probability that a given state-action pair is optimal. The estimated confidence of each state-action pair is utilized as a sample weight in their discriminator loss. As a result, the combined discriminator loss with our state and action encoders is defined as follows:
\begin{equation}
    \begin{aligned}
      \underset{\omega}{\text{max}} \underset{x \sim \mathcal{D}_{\pi}}{\mathbb{E}} \left [ \text{log}\: D_{\omega}(z) \right ] + \underset{(x, y) \: \sim\:  (\tilde{\mathcal{D}}_{O} \cup \tilde{\mathcal{D}}_{N})}{\mathbb{E}} \left [\frac{y}{\epsilon} \text{log} (1 - D_{\omega}(z)) \right ],
   \label{eq:2iwil_gail}
   \end{aligned}
\end{equation}
where $\epsilon$ = $\frac{1}{N_L} \sum_{i=1}^{N_L}y_i$, $z = z^s \oplus z^a$. $z^s$ is a state representation embedded by $SE(s)$, and $z^a$ is an action representation embedded by $AE(a)$.

\subsection{Combined Case $\#$2: Ours + 2IWIL + Manifold Mixup}

For utilizing Manifold mixup (MM), we modeled the per-sample confidence score distribution of $(y_l, \hat{y}_u)$ with a two-component Gaussian mixture model to divide the imperfect demonstrations $\mathcal{D}_I$ into optimal demonstrations and non-optimal demonstrations, $\tilde{\mathcal{D}}_{O}$ and $\tilde{\mathcal{D}}_{N}$. The feature representation $z_{o} \sim \tilde{\mathcal{D}}_{O}$ is interpolated with the feature representation $z_{n} \sim \tilde{\mathcal{D}}_{N}$. More formally, for a batch of features $(z_{o}, z_{n})$ and corresponding confidence scores $(y_{o}, y_{n})$, the mixed $(\bar{z}, \bar{y})$ can be computed by:
\begin{equation}
    \begin{aligned}
      \lambda \sim & \; \text{Beta}(\alpha, \alpha),\;\;\; {\lambda}' = \text{max}(\lambda, 1-\lambda), \\
      & \bar{z} = {\lambda}' \cdot z_{o} + (1 - {\lambda}') \cdot z_{n}, \\
      & \bar{y} = {\lambda}' \cdot y_{o} + (1 - {\lambda}') \cdot y_{n},
   \label{eq:ourmixup}
   \end{aligned}
\end{equation}
where $z = z^s \oplus z^a$. $z^s$ is a state representation embedded by $SE(s)$, $z^a$ is an action representation embedded by $AE(a)$. Eq.~\ref{eq:ourmixup} can ensure that $\bar{z}$ are closer to optimal demonstrations than non-optimal demonstrations. 

As shown in Algorithm~\ref{alg:algo2}, by additionally including synthetic data through MM, the discriminator loss is expressed as follows:
\begin{equation}
    \begin{aligned}
      \underset{\omega}{\text{max}} & \underset{x \sim \mathcal{D}_{\pi}}{\mathbb{E}} \left [ \text{log}\: D_{\omega}(z) \right ] + \underset{(x, y) \: \sim\:  (\tilde{\mathcal{D}}_{O} \cup \tilde{\mathcal{D}}_{N})}{\mathbb{E}} \left [\frac{y}{\epsilon} \text{log} (1 - D_{\omega}(z)) \right ] + \\
      & \; \; \; \; \; \; \; \; \; \; \; \; \; \; \; \; \; \; \; \; \; \; \; \underset{(\bar{z}, \bar{y}) \: \sim\:  (\tilde{\mathcal{D}}_{O} \cup \tilde{\mathcal{D}}_{N})}{\mathbb{E}} [\text{log} ((1-\bar{y}) \cdot D_{\omega}(\bar{z}) + \bar{y} \cdot (1 - D_{\omega}(\bar{z}))) ].
   \label{eq:weightedgail}
   \end{aligned}
\end{equation}


\section{Expert Data Size} \label{sec:appendixd}

We assessed our method with varying expert data sizes. As shown in the table, there is a relatively small or no decrease in the performance up to $N_E$ = 20. When $N_E$ is reduced to 20 for Ant and Walker2d and 10 for HalfCheetah, the performance is comparable to the baseline AIL.

\begin{table}[h]
\centering
\caption{Additional studies using 10, 20, or 50 expert state-action pairs on Ant-v2, HalfCheetah-v2, and Walker2d-v2 of MuJoCo.}
\begin{adjustbox}{width=0.9\linewidth}
\begin{tabular}{cccccc}
\toprule
        & GAIL & \multicolumn{4}{c}{Ours}          \\  \cmidrule(l){1-1} \cmidrule(l){2-2} \cmidrule(l){3-6}
 & $N_E = 100$ & $N_E = 10$ & $N_E = 20$ & $N_E = 50$ & $N_E = 100$  \\
\cmidrule(l){1-1} \cmidrule(l){2-2} \cmidrule(l){3-6}
Ant    &    4198.2$\pm$72.6   & 2855.9$\pm$1139.7     & 4286.4$\pm$125.6  & 4432.4$\pm$41.1 & 4554.8$\pm$162.6   \\
HalfCheetah    &    2034.6$\pm$2384.6   & 2787.6$\pm$3454.4   &5381.6$\pm$81.3   & 5410.6$\pm$44.2  & 5416.0$\pm$203.8  \\
Walker2d    &    3513.4$\pm$172.9  & 3023.1$\pm$439.6   & 3412.6$\pm$195.6  &  3520.9$\pm$108.5 & 3527.6$\pm$131.4  \\
\bottomrule
\end{tabular}
\label{tab:ablationexpert}
\end{adjustbox}
\end{table}






\section{Role of Gaussian Noise} \label{sec:appendixe}

Appending noise dimensions increases the performance on all three tasks as shown in Table~\ref{tab:ablation}. -71.09, -81.29, and –87.90 represent the decrease without appending the noise dimensions. 

\begin{table}[h]
\centering
\caption{Ablation studies on appending Gaussian noise using 100 expert state-action pairs on Ant-v2, HalfCheetah-v2, and  Walker2d-v2 of MuJoCo.}
\begin{adjustbox}{width=.9\linewidth}
\begin{tabular}{cccc}
\toprule
Ours & Ant & HalfCheetah & Walker2d  \\
\cmidrule(l){1-1} \cmidrule(l){2-4}
w/o noise   &    4483.7$\pm$159.6 (-71.09)  & 5334.7$\pm$43.0 (-81.29)     & 3439.7$\pm1$22.2 (-87.90) \\
w noise   &    4554.8$\pm$162.6  & 5416.0$\pm$203.8    & 3527.6$\pm$131.4 \\
\bottomrule
\end{tabular}
\label{tab:ablation}
\end{adjustbox}
\end{table}

\section{Discrete control Benchmarks} \label{sec:appendixf}

To test the scalability of the proposed method, we evaluated the performance of the proposed method on 2 discrete control benchmarks: BeamRider-ram-v0, and SpaceInvaders-ram-v0 of OpenAI Gym. 

\begin{table}[h]
\centering
\caption{Final performance using 20 expert state-action pairs on BeamRider-ram-v0, and SpaceInvaders-ram-v0 of OpenAI Gym. Best results are in \textbf{bold}.}
\begin{adjustbox}{width=.7\linewidth}
\begin{tabular}{cccc}
\toprule
         \multicolumn{2}{c}{BeamRider} & \multicolumn{2}{c}{SpaceInvaders}          \\  \cmidrule(l){1-2} \cmidrule(l){3-4}
  GAIL & Ours & GAIL & Ours  \\
\cmidrule(l){1-2} \cmidrule(l){3-4}
   399.43$\pm$55.63   & \textbf{433.04$\pm$76.15}     & 166.39$\pm$107.88       & \textbf{289.87$\pm$5.37}   \\
\bottomrule
\end{tabular}
\label{tab:discrete}
\end{adjustbox}
\end{table}



For comparison, we chose GAIL, which showed the second-best performance in the experiments of continuous control benchmarks. To apply the proposed method on discrete control benchmarks, we ignored the action encoder $AE$ and its corresponding loss $\mathcal{L}_{AC}$ of the proposed model. Nevertheless, as shown in Table~\ref{tab:discrete}, the proposed method still showed better performance compared to GAIL on discrete control benchmarks.



\end{document}